\documentclass[11pt]{article}

\usepackage[preprint]{acl}

\usepackage{times}
\usepackage{latexsym}

\usepackage[T1]{fontenc}

\usepackage[utf8]{inputenc}

\usepackage{microtype}

\usepackage{inconsolata}

\usepackage{graphicx}

\usepackage{CJKutf8}
\usepackage{newtxmath}
\usepackage{amsmath}
\newcommand{\Dm}{\Delta_{m}}
\usepackage{makecell}
\usepackage{array, multirow}
\usepackage[table, x11names, dvipsnames, xcdraw]{xcolor}
\usepackage[normalem]{ulem}
\useunder{\uline}{\ul}{}
\usepackage{booktabs}
\usepackage{nicematrix}
\usepackage{color, colortbl}
\usepackage{hhline}
\usepackage{subfigure}     

\definecolor{Gray}{gray}{0.9}

\usepackage{enumitem}
\usepackage[super]{nth}
\usepackage{amsfonts}

\usepackage{pmboxdraw}

\usepackage{appendix}

\usepackage{cleveref}
\usepackage{hyperref}

\title{\textsc{Mapes}: Multi-Value Alignment Pipeline via Expert Specialization}

\author{
  \textbf{SungHo Kim}\textsuperscript{1},
  \textbf{SangKeun Lee}\textsuperscript{1,2}
\\
  \textsuperscript{1}Department of Artificial Intelligence, Korea University, Seoul, South Korea \\
  \textsuperscript{2}Department of Computer Science and Engineering, Korea University, Seoul, South Korea \\
\\
  \texttt{\{sungho3268, yalphy\}@korea.ac.kr}
}

\begin{document}

\begin{CJK}{UTF8}{mj}

\maketitle
\begin{abstract}
Aligning large language models (LLMs) with human preferences requires simultaneously satisfying multiple, often competing values. Existing preference optimization methods collapse these into a single implicit signal, and even multi-objective extensions optimize within a shared parameter space, leaving cross-value conflicts unresolved.
We propose \textsc{Mapes}, a multi-value alignment paradigm in which conflict resolution is a first-class concern at every pipeline stage: it first decomposes supervision into value-specific experts trained in isolated parameter subspaces, then recomposes them via preference-aware value merging and a lightweight harmonization stage that aligns the merged model with the expert ensemble.
Across five preference alignment benchmarks, \textsc{Mapes} outperforms strong direct preference optimization (DPO) variants and multi-objective baselines on the benchmark-level average, with no large degradation on GSM8K and GPQA; matched controls isolate the contribution of each stage.\footnote{Our code is available \href{https://github.com/SungHo3268/MAPES}{here}.}
\end{abstract}

\section{Introduction}
\label{sec:introduction}

Human values are inherently multi-faceted~\cite{constitutionalai-2022}.
When we judge whether a response is \emph{good}, we do not rely on a single criterion; rather, we expect it to follow instructions, be honest, provide genuinely helpful content, and avoid misleading claims~\cite{ultrafeedback2024, guo-etal-2024-controllable, zhou-etal-2024-modpo}.
These values are neither independent nor fully compatible; improving one can degrade another, giving rise to \textit{cross-value conflicts}~\cite{jailbroken-2023, guo-etal-2024-controllable}.
Achieving faithful alignment in LLMs, therefore, requires improving multiple values without allowing progress on one dimension to undermine another systematically.

\begin{figure}[t!]
    \centering
    \includegraphics[width=1.\linewidth]{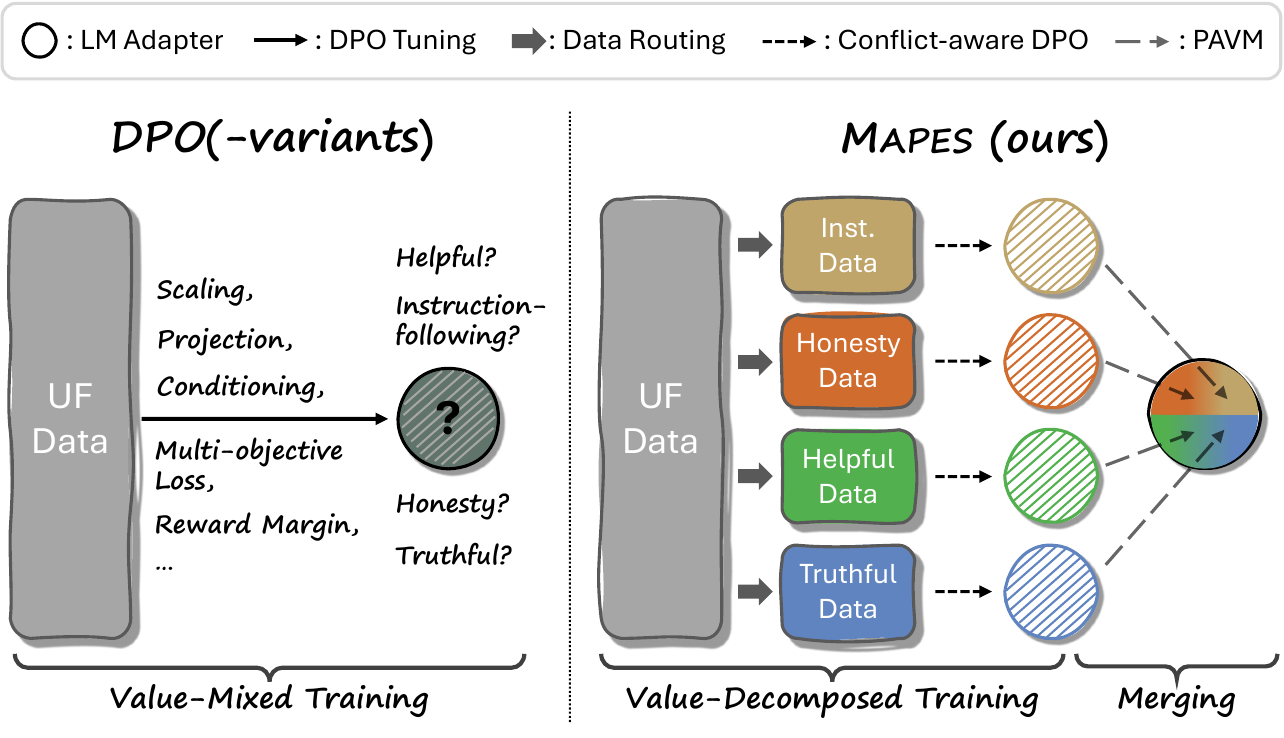}
    \caption{Comparison of existing DPO-based approaches (left) and \textsc{Mapes} (right) for multi-value preference alignment.}
    \label{fig:dpo_mapes_comp}
\end{figure}

To achieve comprehensive multi-value alignment, several recent studies build on DPO~\cite{rafailov2023direct}, which requires no separate reward model. One line of work incorporates target preference scores into training or generation to emphasize a single value at a time within shared parameters~\cite{guo-etal-2024-controllable, yang-2024-rewards, radpo2025}. Another introduces per-value reward margins into a multi-objective DPO objective to balance competing values~\cite{zhou-etal-2024-modpo}. However, both optimize a single shared model, treating cross-value trade-offs as inherent rather than explicitly mitigating them.

Despite these aligning efforts, the structural limit of training within a single shared parameter space remains difficult to fully overcome (left side of Figure~\ref{fig:dpo_mapes_comp}). As different values exert opposing optimization pressures on the same parameters, gains achieved on one value displace, rather than complement, gains on another, a phenomenon long observed for competing objectives in multi-task learning~\citep{sener2018mgda, yu2020pcgrad} and recently shown to exhibit persistent trade-offs across multiple rewards~\citep{rame2023rewardedsoups}, conflicts that added supervision alone cannot resolve.

Mitigating these conflicts suggests a different premise: each value can specialize within its own dedicated parameter subspace, free from competing supervision during training, and only then be carefully composed into a single coherent model. Building on this, we propose \textsc{Mapes} (\textbf{M}ulti-value \textbf{A}lignment \textbf{P}ipeline via \textbf{E}xpert \textbf{S}pecialization), a four-stage pipeline: (i) per-value data routing, (ii) per-value expert training in isolated subspaces, (iii) preference-aware value merging (PAVM), and (iv) post-merge harmonization. The stages target successive layers of cross-value conflict: data, gradient, parameter, and distributional conflicts, respectively. Several components build on established techniques; the contribution lies in their conflict-aware co-design and in PAVM, a new merge rule that weights each expert by the per-coordinate alignment between its delta and its preference-improvement direction.

Empirically, across five preference alignment benchmarks, \textsc{Mapes} attains the highest benchmark-level average for every backbone, although individual columns, mostly instruction-following, can regress. On GSM8K and GPQA we observe no large degradation~\cite{instructgpt-2022}, and harmlessness, a value never used in training, also improves. Matched controls show that ordinary post-merge DPO accounts for most of the recovery after merging, that the ensemble-KL term adds a smaller gain concentrated on honesty, and that the preference direction matters for the merge. Our main contributions are as follows:

\begin{itemize}[leftmargin=*, itemsep=0pt, topsep=2pt]
    \item We recast multi-value alignment as a conflict-aware pipeline problem, where cross-value conflicts are mitigated structurally rather than absorbed into a single shared objective.
    \item We propose \textsc{Mapes}, a systematic four-stage pipeline that addresses successive layers of conflict at the data, gradient, parameter, and distribution levels, with PAVM as its core technique.
    \item We empirically demonstrate the effectiveness of \textsc{Mapes} on five preference alignment benchmarks and isolate the contribution of each stage with matched controls.
\end{itemize}

\section{Related Work}
\label{sec:related_work}

\begin{figure*}[t!]
    \centering
    \includegraphics[width=1.\linewidth]{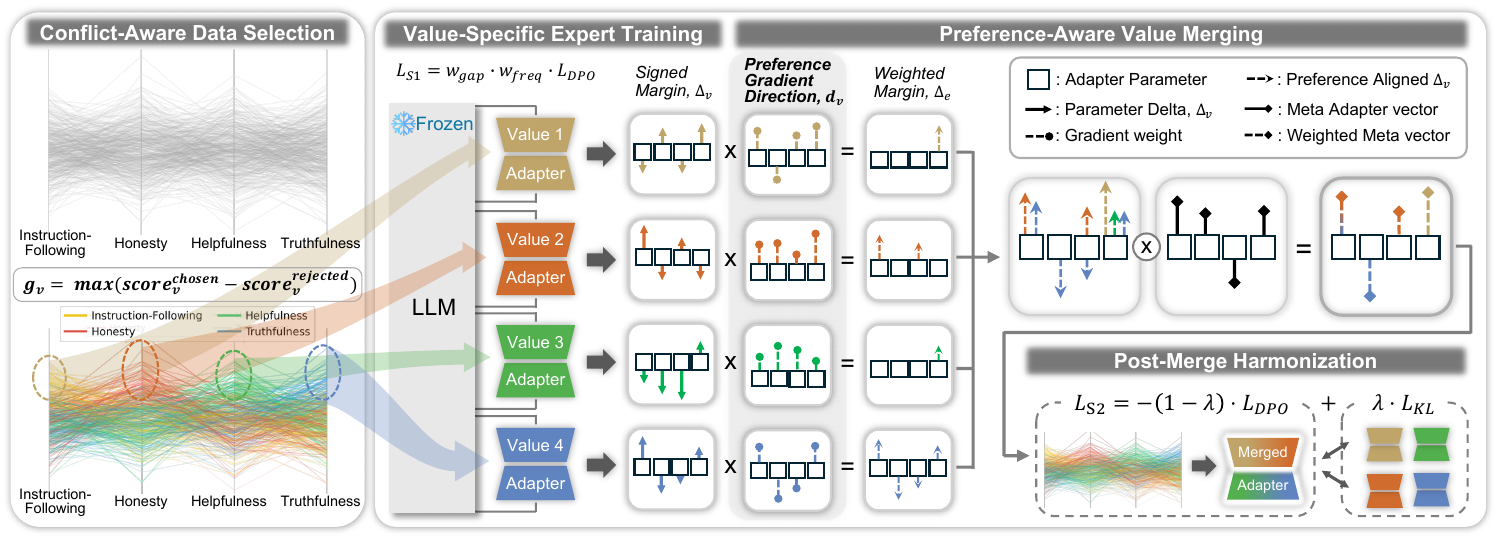}
    \caption{Overview of \textsc{Mapes}. The pipeline first partitions data by value and trains specialized expert adapters via gap-weighted DPO, then unifies the experts into a single adapter through PAVM and post-merge harmonization.}
    \label{fig:mapes_framework}
    \vspace{-1px}
\end{figure*}

\paragraph{Multi-value alignment.}
Existing multi-value alignment methods fall into three groups. (i) Single-model multi-objective methods, such as MODPO~\cite{zhou-etal-2024-modpo}, extend DPO with auxiliary reward margins to balance multiple objectives. (ii) Score-conditioned methods incorporate value scores into direct alignment: RA-DPO~\cite{radpo2025} constructs reward-augmented preference data, while CPO~\cite{guo-etal-2024-controllable} conditions on per-value preference scores. RiC~\cite{yang-2024-rewards}, by contrast, requires separately trained reward models and thus falls outside our reward-model-free setting. (iii) Weight-space methods, such as Rewarded Soups~\cite{rame2023rewardedsoups}, interpolate models trained for individual rewards to approximate a Pareto front. Unlike these approaches, \textsc{Mapes} addresses cross-value conflict across the full pipeline, data, gradient, parameter, and distributional stages, rather than at a single stage. It also requires neither per-value reward models nor frontier sweeping, relying only on standard DPO supervision.

\paragraph{Model merging.}
Model merging combines capabilities from separately fine-tuned models. Task Arithmetic~\cite{ilharco2023editing} adds task vectors, TIES-Merging~\cite{ties2023} mitigates interference through trimming and sign-based selection, and DARE~\cite{yu-et-al-dare-2024} drops delta parameters before aggregation. Designed for task-specialized models, these methods rely mainly on parameter magnitude and do not identify directions that improve preference alignment. In contrast, PAVM (\S\ref{sec:pavm}) incorporates preference-optimization signals and per-value improvement directions relative to the reference model as alignment-aware merge weights, turning magnitude-based aggregation into preference-aware composition.

\section{Methodology}
\label{sec:methodology}

We introduce \textsc{Mapes} (\textbf{M}ulti-value \textbf{A}lignment \textbf{P}ipeline via \textbf{E}xpert \textbf{S}pecialization), a four-stage framework that resolves cross-value conflict throughout the alignment pipeline (Figure~\ref{fig:mapes_framework}): (1)~\emph{data conflict}, score-gap-based assignment routes each sample to its most discriminative value (\S\ref{sec:data_selection}); (2)~\emph{gradient conflict}, value-specific LoRA experts are trained using a gap- and exclusivity-weighted DPO loss (\S\ref{sec:expert_training}); (3)~\emph{parameter conflict}, Preference-Aware Value Merging merges the experts via per-value preference improvement directions (\S\ref{sec:pavm}); and (4)~\emph{distributional conflict}, lightweight harmonization aligns the merged adapter with the expert ensemble (\S\ref{sec:harmonization}).
These stages are co-designed to address residual conflicts from preceding stages. At the base model, per-value preference-improvement gradients are only moderately aligned (mean pairwise cosine $+0.51$) and disagree in sign on approximately $71\%$ of the evaluated coordinates (\S\ref{sec:grad_conflict}); this coordinate-level disagreement is what motivates PAVM's per-coordinate merge rule.

\subsection{Conflict-Aware Data Selection}
\label{sec:data_selection}

We use the UltraFeedback dataset~\cite{ultrafeedback2024}, which provides pairwise preference data with fine-grained value ratings.
Each example is annotated for a set of values $\mathcal{V} = \{\textsc{Instruction-Follow-}\\ \textsc{ing}, \textsc{Honesty}, \textsc{Helpfulness}, \textsc{Truthfulness}\}$ with ordinal scores $s_v^{+}$ and $s_v^{-}$ assigned to the chosen and rejected responses on a 5-point scale.

Given these 36k value-annotated preference data, we first identify a \textit{target value} for each example, defined as the value that most strongly explains the preference between the chosen and rejected responses.
To quantify this, we compute the per-value score gap $g_v = s_v^{+} - s_v^{-}$ for each value $v \in \mathcal{V}$, and assign the example to the value with the largest positive gap, $v^{*} = \arg\max_{v \in \mathcal{V}} g_v$ subject to $g_{v} > 0$.
Examples with no positive gap are discarded, as they do not provide a clear preference signal.

This assignment trains each expert on samples where its target value is most explicitly expressed. After assignment, we balance the number of samples across values via downsampling to 22k preference pairs (details in Appendix~\ref{app:training_dataset}).

\subsection{Value-Specific Expert Training}
\label{sec:expert_training}

\paragraph{Value-specific decomposition.}
For each value $v \in \mathcal{V}$, we train a lightweight LoRA adapter $A_v$ on the corresponding data partition from \S\ref{sec:data_selection}, while keeping the base model $\pi_{\mathrm{base}}$ frozen, yielding a value-specific policy $\pi_{\theta_v} = \pi_{\mathrm{base}} \circ A_v$. Since only the adapter for the assigned value receives gradients for each sample, cross-value interference during expert optimization is minimized by construction; this incurs no additional training cost, as the dataset is partitioned across experts rather than duplicated. To condition each adapter on its target value, we inject the value identity and a target score (on a 1--5 scale) into the system prompt as ``\texttt{[TARGET\_VALUE] score: [TARGET\_SCORE]}'', with the full template provided in Appendix~\ref{app:prompt_exp}.

\paragraph{Gap-Weighted DPO Tuning.}
Data decomposition isolates value signals, but not all samples provide equally strong supervision. We therefore weight each sample by both the strength and the uniqueness of its preference signal:
\begin{equation}
    w = \frac{g_{v^*}}{\mathbb{E}[g_{v^*}]} \cdot \big(1 + \gamma \cdot \max(e, 0)\big),
\end{equation}
where the first factor gives larger preference differences stronger training signals, the exclusivity score $e = g_{v^*} - \bar{g}_{\neg v^*}$ (with $\bar{g}_{\neg v^*}$ the mean gap over non-target values) emphasizes value-specific signals, and $\gamma$ controls the strength of this amplification. Each adapter is then optimized via $\min_{\theta_v} \mathbb{E}\big[ w \cdot \mathcal{L}_{\mathrm{DPO}}(\pi_{\theta_v}) \big]$, where $\mathcal{L}_{\mathrm{DPO}}$ denotes the standard DPO loss~\cite{rafailov2023direct}.

\subsection{Preference-Aware Value Merging}
\label{sec:pavm}

Having trained value-specific experts (\S\ref{sec:expert_training}), we next compose them into a single adapter that simultaneously incorporates all values. Existing model-merging methods, however, are designed for task-specialized models and rely primarily on parameter magnitude, without capturing which directions in parameter space improve \emph{preference alignment} for each value. We propose Preference-Aware Value Merging (PAVM), which, to our knowledge, is the first method to use per-coordinate alignment with DPO-gradient-based preference-improvement directions as merge weights for value-specialized preference adapters. Let $\Delta_v = \theta_v - \theta_{\mathrm{base}}$ denote the parameter delta of expert adapter $A_v$ relative to the base model.

\paragraph{Preference improvement direction.}
For each value $v$, we define a per-value \emph{preference improvement direction} $p_v$ at the base reference $\pi = \pi_{\mathrm{ref}} = \pi_{\mathrm{base}}$ (all expert adapters disabled), where the DPO gradient simplifies to
\begin{equation}
    p_v \;\propto\; \nabla_{\theta} \log \pi_{\mathrm{base}}(y^{+} \mid x) - \nabla_{\theta} \log \pi_{\mathrm{base}}(y^{-} \mid x).
    \label{eq:pref_dir}
\end{equation}
We estimate $p_v$ as the negative gradient of the chosen-vs-rejected NLL gap over a small calibration set ($N{=}128$ preference pairs per value); the full estimation procedure and gradient scope are deferred to Appendix~\ref{app:pavm_details}; PAVM is robust to the trim density $d$, the calibration size $N$, and the direction-estimation reference point (Appendix~\ref{app:pavm_sensitivity}). Intuitively, $p_v$ specifies the direction in parameter space along which moving the base model locally increases preference for value $v$. PAVM then asks: \emph{for each parameter, which expert's learned update is most aligned with its own preference direction?}

\paragraph{Phase 1: Magnitude-Based Trimming.}
To focus on each expert's most influential parameter updates, we retain only the top-$d$ fraction of entries of $\Delta_v$ by magnitude and zero out the rest~\cite{ties2023}, yielding trimmed deltas $\tilde{\Delta}_v$ ($d{=}0.2$).

\paragraph{Phase 2: Sign Election with Preference Tiebreak.}
For each parameter position $j$, signs across experts may disagree. We elect a sign by delta-sum voting over the trimmed deltas, $s_j = \mathrm{sign}\big(\textstyle\sum_{v \in \mathcal{V}} \tilde{\Delta}_{v,j}\big)$.
When the sum is exactly zero (full cancellation across experts), we break the tie using a magnitude-weighted vote over preference directions, $s_j \leftarrow \mathrm{sign}\!\big(\sum_{v} p_{v,j} \cdot |p_{v,j}|\big)$, giving stronger preferences greater influence.

\newcommand{\captioncolorbox}[2]{%
  {\setlength{\fboxsep}{0.5pt}%
   \raisebox{0pt}[0.9em][0pt]{\colorbox[HTML]{#1}{\strut #2}}}%
}

\begin{table*}[t!]
    \centering
    \small
    \setlength{\tabcolsep}{3.5pt} 
    \renewcommand{\arraystretch}{0.90}
    \resizebox{\textwidth}{!}{
    \begin{tabular}{llcccccccccc}
    \toprule
    \multicolumn{1}{c}{\multirow{4}{*}{Model}} & \multicolumn{1}{r}{Value $\rightarrow$} & \multicolumn{3}{c}{Instruction-Following} & \multicolumn{1}{c}{Honesty} & \multicolumn{2}{c}{Helpfulness} & \multicolumn{2}{c}{Truthfulness} & \multicolumn{1}{c}{\multirow{4}{*}{Avg.~$\uparrow$}} & \multicolumn{1}{c}{\multirow{4}{*}{Rank~$\downarrow$}} \\
    \cmidrule(lr){3-5} \cmidrule(lr){6-6} \cmidrule(lr){7-8} \cmidrule(lr){9-10}
    \multicolumn{1}{c}{} & \multicolumn{1}{r}{Benchmark $\rightarrow$} & \multicolumn{2}{c}{Arena-Hard-Auto} & \multicolumn{1}{c}{IFEval} & \multicolumn{2}{c}{HHH-Alignment} & \multicolumn{1}{c}{HH-RLHF} & \multicolumn{2}{c}{TruthfulQA} \\
    \cmidrule(lr){3-4} \cmidrule(lr){5-5} \cmidrule(lr){6-7} \cmidrule(lr){8-8} \cmidrule(lr){9-10}
    \multicolumn{1}{c}{} & \multicolumn{1}{r}{Subset $\rightarrow$} & \multicolumn{1}{r}{v0.1} & v2.0 & Prompt & Honest & Helpful & Helpful& MC1 & MC2 & & \\

    \midrule
    \multicolumn{2}{l}{Qwen2.5-3B-Instruct} & 19.82 & 2.40 & \textbf{35.92} & 44.26 & 38.98 & 60.83 & 24.72 & 41.41 & 36.51{\scriptsize$\pm$0.05} & 5.81 \\
    \multicolumn{2}{l}{Qwen2.5-3B-Instruct + DPO} & 19.64 & 2.22 & 35.30 & 44.26 & 44.07 & 61.21 & 30.35 & 47.45 & 38.10{\scriptsize$\pm$0.35} & 5.31 \\
    \multicolumn{2}{l}{Qwen2.5-3B-Instruct + CDPO} & 21.12 & 1.84 & \textbf{35.92} & 45.36 & 42.37 & 61.04 & 37.33 & 56.24 & 39.82{\scriptsize$\pm$0.33} & 3.88 \\
    \multicolumn{2}{l}{Qwen2.5-3B-Instruct + MODPO} & 19.90 & 2.50 & 35.67 & 45.90 & 40.68 & 62.70 & 33.91 & 49.93 & 38.96{\scriptsize$\pm$0.13} & 3.81 \\
    \multicolumn{2}{l}{Qwen2.5-3B-Instruct + RA-DPO} & \textbf{22.52} & 2.74 & 35.43 & 47.54 & 48.59 & \textbf{62.72} & 27.21 & 47.12 & 39.20{\scriptsize$\pm$0.17} & 3.31 \\
    \multicolumn{2}{l}{Qwen2.5-3B-Instruct + RS} & 21.80 & \textbf{3.20} & 34.57 & 47.54 & 40.68 & 61.55 & 33.41 & 53.09 & 39.20{\scriptsize$\pm$0.18} & 3.50 \\
    \rowcolor[HTML]{EAECFF}
    \multicolumn{2}{l}{Qwen2.5-3B-Instruct + \textsc{Mapes}} & 19.54 & 2.86 & \textbf{35.92} & \textbf{57.38} & \textbf{49.15} & 61.32 & \textbf{47.00} & \textbf{63.38} & \textbf{43.38{\scriptsize$\pm$0.27}}$^{**}$ & \textbf{2.38} \\

    \midrule
    \multicolumn{2}{l}{Qwen2.5-7B-Instruct} & 29.80 & 3.72 & 39.74 & 49.18 & 38.98 & 62.66 & 33.66 & 51.64 & 41.18{\scriptsize$\pm$0.06} & 5.75 \\
    \multicolumn{2}{l}{Qwen2.5-7B-Instruct + DPO} & 31.14 & 3.54 & 39.25 & 54.10 & 42.37 & 62.53 & 34.80 & 53.25 & 42.28{\scriptsize$\pm$0.48} & 5.50 \\
    \multicolumn{2}{l}{Qwen2.5-7B-Instruct + CDPO} & 31.42 & 4.26 & 38.32 & 55.74 & 52.54 & 61.12 & 38.92 & 60.34 & 44.21{\scriptsize$\pm$0.43} & 4.25 \\
    \multicolumn{2}{l}{Qwen2.5-7B-Instruct + MODPO} & 29.60 & 4.00 & \textbf{40.48} & 50.82 & 44.07 & 62.34 & 36.84 & 55.56 & 42.65{\scriptsize$\pm$0.46} & 5.00 \\
    \multicolumn{2}{l}{Qwen2.5-7B-Instruct + RA-DPO} & 31.54 & 4.82 & 39.50 & 62.30 & 58.76 & 62.41 & 38.84 & 60.71 & 46.08{\scriptsize$\pm$0.33} & 3.12 \\
    \multicolumn{2}{l}{Qwen2.5-7B-Instruct + RS} & 31.20 & 4.80 & 34.94 & 67.21 & 59.32 & 63.04 & \textbf{45.04} & 60.03 & 46.36{\scriptsize$\pm$0.21} & 3.12 \\
    \rowcolor[HTML]{EAECFF}
    \multicolumn{2}{l}{Qwen2.5-7B-Instruct + \textsc{Mapes}} & \textbf{34.20} & \textbf{5.28} & 39.93 & \textbf{72.13} & \textbf{67.80} & \textbf{63.22} & 43.33 & \textbf{61.15} & \textbf{49.02{\scriptsize$\pm$0.21}}$^{**}$ & \textbf{1.25} \\

    \midrule
    \multicolumn{2}{l}{Llama-3.2-3B-Instruct} & 10.90 & 1.02 & 36.60 & 42.62 & 38.98 & 53.65 & 24.24 & 46.32 & 34.46{\scriptsize$\pm$0.10} & 5.75 \\
    \multicolumn{2}{l}{Llama-3.2-3B-Instruct + DPO} & 11.24 & 1.10 & 36.04 & 49.18 & 40.68 & 61.82 & 37.13 & 57.51 & 39.26{\scriptsize$\pm$0.09} & 4.69 \\
    \multicolumn{2}{l}{Llama-3.2-3B-Instruct + CDPO} & 10.24 & 0.84 & 35.31 & 57.38 & 52.54 & 62.61 & 40.31 & 60.05 & 41.72{\scriptsize$\pm$0.26} & 3.75 \\
    \multicolumn{2}{l}{Llama-3.2-3B-Instruct + MODPO} & \textbf{12.90} & 0.80 & \textbf{36.66} & 45.90 & 44.07 & 61.38 & 39.29 & 58.00 & 39.70{\scriptsize$\pm$0.41} & 4.44 \\
    \multicolumn{2}{l}{Llama-3.2-3B-Instruct + RA-DPO} & 9.02 & 0.94 & 36.41 & 59.02 & 49.15 & 62.33 & 43.13 & 60.72 & 41.95{\scriptsize$\pm$0.36} & 3.38 \\
    \multicolumn{2}{l}{Llama-3.2-3B-Instruct + RS} & 11.10 & 1.10 & 33.27 & 50.82 & 44.07 & \textbf{64.10} & 40.27 & 59.63 & 40.17{\scriptsize$\pm$0.19} & 3.88 \\
    \rowcolor[HTML]{EAECFF}
    \multicolumn{2}{l}{Llama-3.2-3B-Instruct + \textsc{Mapes}} & 11.64 & \textbf{1.32} & 34.38 & \textbf{62.30} & \textbf{62.71} & 62.32 & \textbf{44.80} & \textbf{61.02} & \textbf{43.72{\scriptsize$\pm$0.21}}$^{**}$ & \textbf{2.12} \\

    \midrule
    \multicolumn{2}{l}{Llama-3.1-8B-Instruct} & 15.22 & 1.64 & 28.53 & 44.26 & 38.98 & 55.18 & 26.32 & 48.51 & 34.24{\scriptsize$\pm$0.05} & 6.75 \\
    \multicolumn{2}{l}{Llama-3.1-8B-Instruct + DPO} & 16.94 & 1.68 & 33.64 & 62.30 & 55.93 & 63.20 & 39.49 & 58.54 & 42.86{\scriptsize$\pm$0.42} & 4.44 \\
    \multicolumn{2}{l}{Llama-3.1-8B-Instruct + CDPO} & 18.16 & 1.72 & 33.33 & 60.66 & 57.63 & 65.04 & 46.02 & 62.23 & 44.32{\scriptsize$\pm$0.73} & 3.50 \\
    \multicolumn{2}{l}{Llama-3.1-8B-Instruct + MODPO} & 18.20 & 1.40 & 36.60 & 47.54 & 48.59 & 63.51 & 38.19 & 56.91 & 41.10{\scriptsize$\pm$0.13} & 5.12 \\
    \multicolumn{2}{l}{Llama-3.1-8B-Instruct + RA-DPO} & 16.64 & 1.42 & 34.20 & 59.02 & 55.93 & \textbf{65.22} & 45.00 & 62.24 & 43.91{\scriptsize$\pm$0.15} & 4.06 \\
    \multicolumn{2}{l}{Llama-3.1-8B-Instruct + RS} & \textbf{20.60} & \textbf{2.40} & 35.67 & 59.02 & 52.54 & 64.40 & \textbf{46.63} & 63.47 & 44.48{\scriptsize$\pm$0.20} & 2.69 \\
    \rowcolor[HTML]{EAECFF}
    \multicolumn{2}{l}{Llama-3.1-8B-Instruct + \textsc{Mapes}} & 19.24 & 2.16 & \textbf{37.03} & \textbf{63.93} & \textbf{61.02} & \textbf{65.22} & 46.43 & \textbf{64.72} & \textbf{46.20{\scriptsize$\pm$0.51}}$^{**}$ & \textbf{1.44} \\
    \bottomrule
    \end{tabular}
    }
    \caption{Evaluation across five preference alignment benchmarks (Arena-Hard: LLM-judged win rate; IFEval: prompt-level strict accuracy; HHH/HH-RLHF: log-likelihood preference accuracy; TruthfulQA: MC1/MC2). \textsc{Mapes} is shaded \captioncolorbox{EAECFF}{blue}; best per family in \textbf{bold}. Avg: benchmark-level macro average over the five benchmarks (the two columns of Arena-Hard, HHH-Alignment and TruthfulQA are averaged first), mean{\scriptsize$\pm$}std over three seeds (per-cell deviations in Appendix~\ref{app:seed_std}); Rank: mean rank across the eight scored columns; $^{**}$: significant Avg gain over the strongest DPO-based baseline (paired $t$-test over seeds, $p<0.05$).}
    \label{tab:main_results}
    \vspace{-5px}
\end{table*}

\paragraph{Phase 3: Preference-Aligned Aggregation.}
This is the core of PAVM. For each expert $v$, we measure the element-wise alignment between its learned delta and its own preference improvement direction, $a_{v,j} = \max\!\big(\Delta_{v,j} \cdot p_{v,j},\, 0\big)$, keeping only positively-aligned components, since a delta that moves \emph{against} the preference direction provides no preference-improving evidence regardless of its magnitude. We normalize alignment \emph{per coordinate} across experts, $\tilde{a}_{v,j} = a_{v,j} / \max_{v'\in\mathcal{V}} a_{v',j}$, so that at each parameter position the most preference-aligned expert receives full credit and the others are scaled relative to this local maximum, yielding $\tilde{a}_{v,j} \in [0, 1]$. The preference-aligned merge weight $w_{v,j} = 1 + \alpha \cdot \tilde{a}_{v,j}$ then gives higher merge influence to experts whose deltas are most positively aligned with their own preference direction ($w_{v,j} \in [1, 1{+}\alpha]$, with $\alpha{=}1.0$ by default). The merged delta aggregates the trimmed deltas of the experts whose signs agree with the elected sign $s_j$, weighted by $w_{v,j}$:
\begin{equation}
    \Delta_{\mathrm{merged},\,j} \;=\; \frac{\sum_{v \in \mathcal{V}_j^{\star}} w_{v,j} \cdot \tilde{\Delta}_{v,j}}{\max\!\big(\sum_{v \in \mathcal{V}_j^{\star}} w_{v,j},\;1\big)},
    \label{eq:pavm_merge}
\end{equation}
where $\mathcal{V}_j^{\star} = \{v : \tilde{\Delta}_{v,j} \neq 0,\ \mathrm{sign}(\tilde{\Delta}_{v,j}) = s_j\}$; the alignment $a_{v,j}$ is computed on the untrimmed delta, which coincides with $\tilde{\Delta}_{v,j}$ wherever the latter is nonzero.\footnote{The denominator clamping handles the corner case in which no expert agrees with the elected sign at position $j$ ($\mathcal{V}_j^{\star} = \emptyset$); $\Delta_{\mathrm{merged},j}$ is then set to zero.}

\paragraph{Local properties.}
Eq.~(\ref{eq:pavm_merge}) has three local properties, proved in Appendix~\ref{app:pavm_theory}. (i)~At each coordinate it is the $w_{v,j}$-weighted least-squares barycenter of the sign-consistent expert deltas. (ii)~It decomposes exactly as $\Delta_{\mathrm{merged},j} = \Delta^{\mathrm{unif}}_{j} + (\alpha N_j / W_j)\,\mathrm{Cov}_v(\tilde{a}_{v,j}, \tilde{\Delta}_{v,j})$, where $\Delta^{\mathrm{unif}}_j$ is the uniform (TIES-style, $\alpha{=}0$) aggregate, $N_j = |\mathcal{V}_j^{\star}|$ and $W_j = \sum_{v\in\mathcal{V}_j^{\star}} w_{v,j}$; in particular the preference weighting never flips the elected sign, and $\|\Delta_{\mathrm{merged}}\| \le (\sum_v \|\tilde{\Delta}_v\|^2)^{1/2}$, so the bounded-step assumption of Proposition~1 holds automatically. (iii)~Whenever alignment scores and delta magnitudes are similarly ordered across $\mathcal{V}_j^{\star}$, $|\Delta_{\mathrm{merged},j}| \ge |\Delta^{\mathrm{unif}}_j|$, so the first-order aggregate gain $\sum_v \langle p_v, \Delta_{\mathrm{merged}}\rangle$ (a controlled surrogate of the true loss change under local smoothness, Proposition~1) is at least uniform's on coordinates whose elected sign agrees with $\sum_v p_{v,j}$. These are local, conditional statements, not guarantees; the $\alpha{=}0$ and random-direction rows of \S\ref{sec:ablation} test them.

\subsection{Post-Merge Harmonization}
\label{sec:harmonization}

PAVM resolves parameter-level conflicts, but composing parameters from distinct optimization trajectories can yield a distribution structurally inconsistent with any of the experts that produced it, a \emph{distributional conflict} we empirically confirm in \S\ref{sec:signal_maintenace}. We address this with a lightweight harmonization stage that combines value-conditioned DPO on the merged adapter with a forward KL distillation from the expert ensemble distribution $\bar{\pi}_{\mathrm{expert}}$, defined as the logit average across the value-specific experts:
\begin{equation}
    \mathcal{L}_{\mathrm{harm}} \;=\; \mathcal{L}_{\mathrm{DPO}}(\pi_{\theta_{\mathrm{merged}}}) \;+\; \lambda \cdot D_{\mathrm{KL}}\!\big(\bar{\pi}_{\mathrm{expert}} \;\big\|\; \pi_{\theta_{\mathrm{merged}}}\big).
    \label{eq:harm}
\end{equation}

The DPO term maintains preference alignment under the unified value condition; the KL term anchors the merged adapter to the experts' collective behavior. A $\lambda{=}0$ control from the same merged checkpoint separates the two terms (\S\ref{sec:ablation}). Harmonization fits within a training budget comparable to a standard DPO baseline at matched optimization steps, reallocating updates from simple amplification toward distribution-level coherence; in wall-clock terms, the full \textsc{Mapes} pipeline requires approximately 15--19\% more GPU-hours than standard DPO (Appendix~\ref{app:compute_cost}).

\section{Experiments}
\label{sec:experiments}
We evaluate \textsc{Mapes} on multi-value preference alignment (\S\ref{sec:main_results}), positive transfer to an unseen value (harmlessness) (\S\ref{sec:positive_transfer}), and alignment-tax effects on general academic reasoning (\S\ref{sec:academic_evaluation}).

\subsection{Experimental Settings}
\paragraph{Baselines.}
We compare \textsc{Mapes} with strong offline preference optimization baselines: standard DPO~\cite{rafailov2023direct}; CPO~\cite{guo-etal-2024-controllable} (value-specific rating tokens; trained with the CDPO strategy); RA-DPO~\cite{radpo2025} (reward-augmented preference data); and MODPO~\cite{zhou-etal-2024-modpo}, a multi-objective extension of DPO. All methods share the same instruction-tuned base models, Qwen2.5-3B/7B~\cite{qwen2025qwen25technicalreport}, Llama-3.2-3B~\cite{meta2024llama32blog}, and Llama-3.1-8B~\cite{dubey2024llama3herdmodels}, and a unified reference model; per-method configurations are in Appendix~\ref{app:training_details}. We additionally report RS, a Rewarded-Soups-style~\cite{rame2023rewardedsoups} baseline that uniformly interpolates our value expert modules with no post-merge training; it therefore uses three quarters of the update budget of the other methods (Appendix~\ref{app:rs_sweep} reports non-uniform weights). Code: \url{https://anonymous.4open.science/r/MAPES-F34F/}.

\paragraph{Tasks.}
We evaluate alignment across multiple human values: instruction-following with Arena-Hard-Auto~\cite{arena-hard-2024} and IFEval~\cite{zhou2023instruction}; helpfulness and honesty with HHH-Alignment~\cite{hhh-2021} and HH-RLHF~\cite{hh-rlhf-2022}; truthfulness with TruthfulQA~\cite{lin-etal-2022-truthfulqa}. The harmlessness subsets of HHH-Alignment and HH-RLHF test positive transfer to an unseen value, and GSM8K~\cite{gsmk8k-2021} (0-shot CoT generation) and GPQA~\cite{gpqa-2023} (log-likelihood multiple choice) check alignment-tax effects.

\paragraph{Training details.}
All models are trained on the binarized UltraFeedback dataset~\cite{ultrafeedback2024, zephyr-2023}: after filtering, approximately 36k preference pairs, which \textsc{Mapes} partitions by value and balances to approximately 22k samples (Appendix~\ref{app:data_preprocessing}). We fix total update steps to four epochs of standard DPO; \textsc{Mapes} allocates three epochs to expert training and one to harmonization, matching the baselines' update count since the data are partitioned, not replicated, across experts. All reported numbers are averaged over three random seeds (42/43/44). For every method, including \textsc{Mapes}, the learning rate is selected from $\{1\mathrm{e}{-6}, 5\mathrm{e}{-6}, 1\mathrm{e}{-5}\}$ by validation loss on a 2,000-pair held-out split of the training data; no evaluation benchmark is used for model or hyperparameter selection. All hyperparameters, including \textsc{Mapes}-specific values and the selected per-backbone learning rates, are consolidated in Appendix~\ref{app:training_details}; Appendix~\ref{app:recipe} gives the end-to-end recipe. At inference time, each method is evaluated under its own native conditioning prompt (full templates in Appendix~\ref{app:prompt_exp}); a control experiment in Appendix~\ref{app:prompt_control} verifies that \textsc{Mapes}'s gains do not stem from its inference prompt.

\subsection{Main Results}
\label{sec:main_results}
Table~\ref{tab:main_results} presents results across the five preference alignment benchmarks (eight scored columns). On the benchmark-level average, \textsc{Mapes} ranks first for every backbone and achieves the best mean rank (overall rank 1.80 vs.\ 3.30). It improves by 3.3--6.7 points over standard DPO and by 1.7--3.6 points over the strongest competing baseline, with the smallest margin on the largest backbone, Llama-3.1-8B (+1.72). RS, a uniform soup of the same experts with no post-merge training, is competitive with the strongest score-conditioned baselines and trails \textsc{Mapes} by 1.7--4.2 points. The average hides regressions on individual columns: \textsc{Mapes} is below the base model on Llama-3.2-3B IFEval (36.60 vs.\ 34.38) and on Qwen2.5-3B Arena-Hard v0.1 (19.82 vs.\ 19.54), and near-tied cells should be read as comparable rather than strict improvements. MODPO, despite explicitly targeting multi-objective preference optimization, underperforms \textsc{Mapes} at every scale.
Across individual value dimensions, the largest and most consistent gains are in honesty and truthfulness, two dimensions known to be difficult to improve simultaneously~\cite{instructgpt-2022, lin-etal-2022-truthfulqa}, while instruction-following is where regressions occur.

\subsection{Evaluation on an Untrained Value}
\label{sec:positive_transfer}

\begin{table}[h!]
    \centering
    \renewcommand{\arraystretch}{0.95}
    \small
    \resizebox{\columnwidth}{!}{
    \begin{tabular}{lccr@{\hspace{6pt}}l}
    \toprule
    \multicolumn{1}{c}{\multirow{2.5}{*}{Model}} & \multicolumn{2}{c}{Harmlessness} & \multicolumn{2}{c}{\multirow{2.5}{*}{Avg.\ {\scriptsize\textcolor{gray}{($+\Delta$)}}}} \\
    \cmidrule(lr){2-3}
    \multicolumn{1}{c}{} & HHH-Align. & HH-RLHF & & \\
    \midrule
    Qwen2.5-7B & 32.76 & 41.18 & 36.97 & \\
    + DPO & 39.66 & 43.08 & 41.37 & {\scriptsize\textcolor{gray}{(+4.4)}}\\
    + CDPO & 44.83 & 43.81 & 44.32 & {\scriptsize\textcolor{gray}{(+7.4)}}\\
    + MODPO & 39.66 & 42.82 & 41.24 & {\scriptsize\textcolor{gray}{(+4.3)}}\\
    + RA-DPO & 62.07 & 43.60 & 52.84 & {\scriptsize\textcolor{gray}{(+15.9)}}\\
    + RS & 56.90 & 43.21 & 50.06 & {\scriptsize\textcolor{gray}{(+13.1)}}\\
    \rowcolor[HTML]{EAECFF}
    + \textsc{Mapes} & \textbf{72.41} & \textbf{43.64} & \textbf{58.03} & {\scriptsize\textcolor{gray}{\textbf{(+21.1)}}}\\
    \midrule
    Llama-3.1-8B & 41.38 & 50.39 & 45.89 & \\
    + DPO & 48.62 & 45.72 & 47.17 & {\scriptsize\textcolor{gray}{(+1.3)}}\\
    + CDPO & 50.00 & 47.49 & 48.75 & {\scriptsize\textcolor{gray}{(+2.9)}}\\
    + MODPO & 49.70 & 42.10 & 45.90 & {\scriptsize\textcolor{gray}{(+0.0)}}\\
    + RA-DPO & 55.17 & 47.84 & 51.51 & {\scriptsize\textcolor{gray}{(+5.6)}}\\
    + RS & 48.28 & 45.93 & 47.11 & {\scriptsize\textcolor{gray}{(+1.2)}}\\
    \rowcolor[HTML]{EAECFF}
    + \textsc{Mapes} & \textbf{56.90} & \textbf{50.69} & \textbf{53.80} & {\scriptsize\textcolor{gray}{\textbf{(+7.9)}}}\\
    \bottomrule
    \end{tabular}
    }
    \caption{Preference evaluation on harmlessness, a value not used in training. Gray parentheses: gain in Avg.\ over the base model; \textsc{Mapes} achieves the largest gain on both backbones.}
    \label{tab:harmless_results}
\end{table}

We also evaluate on harmlessness, a value that is never used as a training target. As shown in Table~\ref{tab:harmless_results}, \textsc{Mapes} improves harmlessness more than all baselines on both backbones. We report this as an evaluation on an untrained value, not as evidence of a transfer mechanism: harmlessness is correlated with the trained values in UltraFeedback, so any method may carry learned behavior over to it; Table~\ref{tab:harmless_results} shows only that \textsc{Mapes} does so the most.

\begin{table*}[t!]
    \centering
    \renewcommand{\arraystretch}{0.95}
    \small
    \setlength{\tabcolsep}{3.5pt}
    \renewcommand{\arraystretch}{0.95}
    \resizebox{\textwidth}{!}{
    \begin{tabular}{lccccccccc}
    \toprule
    \multicolumn{1}{c}{\multirow{4}{*}{Ablation}} & \multicolumn{3}{c}{Instruction-Following} & \multicolumn{1}{c}{Honesty} & \multicolumn{2}{c}{Helpfulness} & \multicolumn{2}{c}{Truthfulness} & \multicolumn{1}{c}{\multirow{4}{*}{Avg.}} \\
    \cmidrule(lr){2-4} \cmidrule(lr){5-5} \cmidrule(lr){6-7} \cmidrule(lr){8-9} & \multicolumn{2}{c}{Arena-Hard} & \multicolumn{1}{c}{IFEval} & \multicolumn{2}{c}{HHH-Align.} & \multicolumn{1}{c}{HH-RLHF} & \multicolumn{2}{c}{TruthfulQA} \\
    \cmidrule(lr){2-3} \cmidrule(lr){4-4} \cmidrule(lr){5-6} \cmidrule(lr){7-7} \cmidrule(lr){8-9} & v0.1 & v2.0 & Prompt & Honest & Helpful & Helpful & MC1 & MC2 & \\
    \midrule
    \textit{Qwen2.5-7B-Instruct} & 29.80 & 3.72 & 39.74 & 49.18 & 38.98 & 62.66 & 33.66 & 51.64 & 41.18 \\
    \rowcolor[HTML]{EAECFF}
    \textbf{\quad + \textsc{Mapes} (Full Pipeline)} & \textbf{34.20} & \textbf{5.28} & \textbf{39.93} & \textbf{72.13} & \textbf{67.80} & \textbf{63.22} & \textbf{43.33} & \textbf{61.15} & \textbf{49.02} \\

    \midrule
    \rowcolor{gray!15} \multicolumn{10}{c}{\textit{Data Selection \& Individual Expert Training}} \\
    \midrule
    Score-Gap Sampling $\rightarrow$ Random Sampling & 24.70 & 3.20 & 34.38 & 68.85 & 59.32 & 61.98 & 43.00 & 60.74 & 45.25 \\
    Multi Experts $\rightarrow$ Single Expert & 30.50 & 3.80 & 36.41 & 72.13 & 55.93 & 61.94 & 42.96 & 58.37 & 46.04 \\
    Gap-Weighted DPO $\rightarrow$ Plain DPO & 26.60 & 3.00 & 38.45 & 70.49 & 67.80 & 62.06 & 42.96 & 60.12 & 47.20 \\

    \midrule
    \rowcolor{gray!15} \multicolumn{10}{c}{\textit{Expert Merging Method}} \\
    \midrule
    PAVM $\rightarrow$ Task Arithmetic~\cite{ilharco2023editing} & 31.90 & 3.40 & 38.08 & 72.13 & 64.41 & 61.51 & 36.84 & 55.70 & 46.36 \\
    PAVM $\rightarrow$ TIES~\cite{ties2023} (= PAVM with $\alpha{=}0$, same implementation) & 30.80 & 4.80 & 38.08 & 72.13 & 67.80 & 62.11 & 40.06 & 56.93 & 47.29 \\
    PAVM $\rightarrow$ DARE(TIES)~\cite{yu-et-al-dare-2024} & 28.70 & 3.90 & 33.64 & 70.49 & 64.41 & 61.43 & 36.11 & 55.42 & 44.92 \\
    PAVM $\rightarrow$ PAVM with random $p_v$ & 23.38 & 3.25 & 32.04 & 70.13 & 65.50 & 61.77 & 41.20 & 55.71 & 44.68 \\

    \midrule
    \rowcolor{gray!15} \multicolumn{10}{c}{\textit{Post-Merge Harmonization}} \\
    \midrule
    Full Pipeline $\rightarrow$ w/o Harmonization (PAVM merge only) & 0.30 & 0.00 & 19.96 & 52.46 & 45.76 & 63.30 & 36.47 & 55.46 & 35.70 \\
    Full Pipeline $\rightarrow$ DPO-only harmonization ($\lambda{=}0$, same merged checkpoint and steps) & 24.90 & 3.05 & 34.01 & 68.85 & 69.49 & 61.64 & 44.80 & 61.50 & 46.39 \\
    \bottomrule
    \end{tabular}
    }
    \caption{Ablation study on Qwen2.5-7B; the \captioncolorbox{EAECFF}{blue row} is the full pipeline. Each row replaces one component (left of $\rightarrow$) with an alternative (right). The random-$p_v$ and $\lambda{=}0$ rows use the identical merge implementation and harmonization schedule (Appendix~\ref{app:harm_controls}).
    }
    \label{tab:ablation_results}
    \vspace{-7px}
\end{table*}

\subsection{Academic Benchmark Evaluation}
\label{sec:academic_evaluation}

\begin{figure}[h!]
    \centering
    \includegraphics[width=\linewidth]{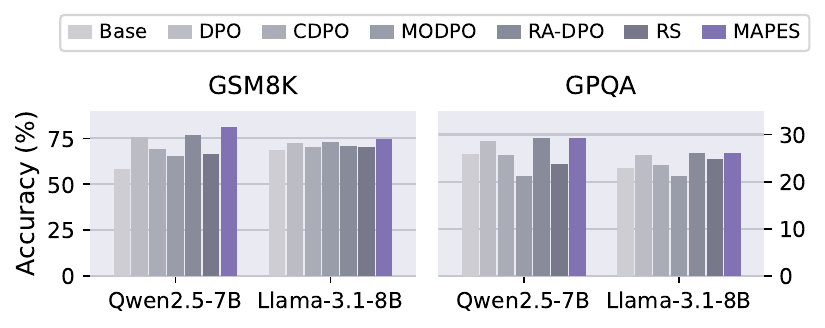}
    \caption{Evaluation results on two academic benchmarks. \textsc{Mapes} shows no large degradation relative to the base model on either test; exact numbers for all four backbones are in Appendix~\ref{app:academic_3b}.}
    \label{fig:academic_bars}
\end{figure}

Preference alignment is often suspected to incur an alignment tax, where gains in alignment come at the expense of general reasoning performance. To investigate this, we evaluate on two academic reasoning benchmarks, namely GSM8K and GPQA. As shown in Figure~\ref{fig:academic_bars}, existing alignment baselines perform comparably to, or slightly better than, the base model. \textsc{Mapes} shows no large degradation on either test and is above the base model on the 7B/8B backbones; two tests cannot establish preserved reasoning in general, and we do not claim it. The unusually large GSM8K gain on Qwen2.5-7B is partly attributable to improved answer-format compliance under the 0-shot protocol rather than stronger underlying reasoning. On the 3B backbones \textsc{Mapes} stays within about one point of the base model on GSM8K and is slightly below the base average on Qwen2.5-3B (39.99 vs.\ 40.29; Appendix~\ref{app:academic_3b}). Overall, we observe no large alignment tax on these two tests.

\section{Analysis}
\label{sec:analysis}

\subsection{Value Gradient Conflict}
\label{sec:grad_conflict}

Appendix~\ref{app:grad_geometry} reports the pairwise cosine similarity between per-value preference-improvement gradients $p_v$, measured at the base model exactly as in PAVM's direction estimation (\S\ref{sec:pavm}; Qwen2.5-7B, $N{=}128$ calibration pairs per value, attention-projection coordinates). The value gradients are moderately aligned, averaging $+0.51$ across pairs (weakest: helpfulness vs.\ instruction-following, $+0.39$), and at least one of the four gradients disagrees in sign with the others on approximately $71\%$ of the evaluated coordinates. These statistics describe gradient geometry; they are not evidence that improving one value must hurt another, since with all cosines positive a common first-order improvement direction exists in the measured subspace. What the sign disagreement does show is that one scalar merge weight per expert is a poor fit at most coordinates, which PAVM's per-coordinate rule addresses; the behavioral evidence for isolation and for the merge rule is \S\ref{sec:ablation}.

\subsection{Ablation Study}
\label{sec:ablation}

Table~\ref{tab:ablation_results} isolates each component's effect by replacing it while holding the rest of the pipeline fixed. Random value assignment in place of score-gap routing degrades the average substantially, and replacing the value-specific experts with a single expert trained on the same routed data lowers the average from 49.02 to 46.04. We read the latter as evidence for the utility of expert isolation, not as evidence of gradient conflict (\S\ref{sec:grad_conflict}).

Among merge operators, Task Arithmetic~\cite{ilharco2023editing}, TIES-Merging~\cite{ties2023}, and DARE~\cite{yu-et-al-dare-2024} all underperform PAVM. The TIES row is PAVM with $\alpha{=}0$ under the identical implementation, so it is the direct test of the preference weighting (47.29 vs.\ 49.02, with the gap concentrated on Arena-Hard and TruthfulQA); replacing $p_v$ by random directions lowers IFEval, honesty, TruthfulQA and Arena-Hard (44.68 vs.\ 49.02 on the average; Table~\ref{tab:ablation_results}), so the direction itself, not only the weighting scheme, matters.

Finally, the bottom block separates the two ingredients of Stage~4. Evaluating the PAVM-merged adapter \emph{before} harmonization drops IFEval and the HHH subsets by ${\sim}20$ points and collapses the Arena-Hard win rate, mirroring the post-merge KL spike in Figure~\ref{fig:kl_trajectory} (right). Continuing the same merged checkpoint with DPO only ($\lambda{=}0$, matched steps) already recovers most of this (46.39 vs.\ 49.02 on the average for the full pipeline); the ensemble-KL term improves honesty by 3.3 points and Arena-Hard v0.1/v2.0 by 9.3/2.2 points relative to DPO-only harmonization, increasing the five-benchmark macro average from 46.39 to 49.02 (+2.6 points), while neither TruthfulQA column improves. Post-merge training, rather than the KL term specifically, accounts for most of Stage~4's effect (full table in Appendix~\ref{app:harm_controls}).

\subsection{Preference Signal Transition}
\label{sec:signal_maintenace}

\begin{figure}[h!]
    \centering
    \includegraphics[width=1.\linewidth]{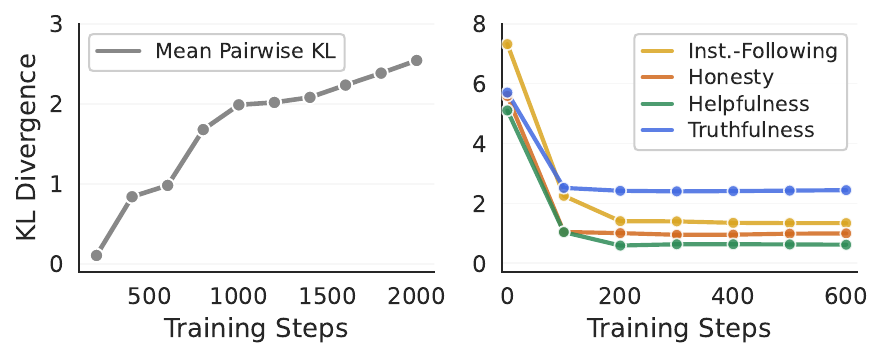}
    \caption{KL divergence trajectories during expert specialization (left) and post-merge harmonization (right): experts diverge during independent training, while the merged model rapidly converges toward all experts.}
    \label{fig:kl_trajectory}
\end{figure}

In Figure~\ref{fig:kl_trajectory} (left), pairwise KL divergence between value-specific experts steadily increases during training, indicating progressive value specialization: experts optimized for different preference signals occupy increasingly distinct regions of the output distribution space.

The right panel shows two phases of post-merge dynamics: immediately after PAVM, the merged adapter exhibits a sharp rise in KL divergence from all experts, indicating that parameter-level merging alone produces a distribution inconsistent with the constituent experts; harmonization then realigns it within a few hundred steps. This is evidence of distributional convergence only; whether value-specific behavior is retained is a behavioral question, addressed by the ceiling comparison in \S\ref{sec:pavm_ceiling_comp} and the controls in \S\ref{sec:ablation}.

\subsection{\textsc{Mapes} Ceiling Performance}
\label{sec:pavm_ceiling_comp}

\begin{figure}[h!]
    \centering
    \includegraphics[width=.80\linewidth]{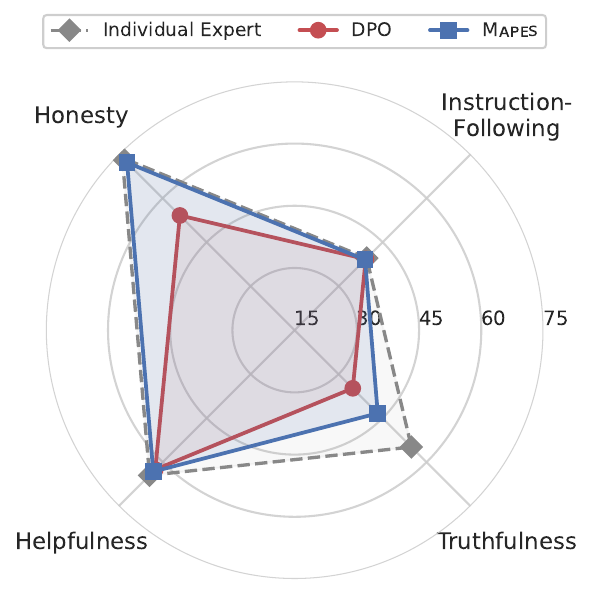}
    \caption{Ceiling performance of value-specific experts vs.\ the merged model. Each expert upper-bounds its target value.}
    \label{fig:ceiling_performance}
\end{figure}

Figure~\ref{fig:ceiling_performance} compares the ceiling performance of value-specific experts with that of the merged \textsc{Mapes} adapter. Each expert achieves a clear peak on its target value and substantially outperforms standard DPO, confirming effective specialization from conflict-aware data selection and expert training. The merged model preserves most of this expert-level performance and nearly reaches the expert-defined ceiling for instruction-following, honesty, and helpfulness, indicating that PAVM and harmonization effectively integrate the experts; a visible gap remains for truthfulness, suggesting residual conflict and room for improvement.

\section{Conclusion}
\label{sec:conclusion}
We propose \textsc{Mapes}, a multi-value alignment pipeline that treats conflict resolution as a first-class concern across all stages of the pipeline, data, gradient, parameter, and distribution, rather than isolating it within a single component. To our knowledge, PAVM is the first method to use per-coordinate alignment with DPO-gradient-based preference-improvement directions as merge weights for value-specialized preference adapters. It is co-designed with a harmonization stage that resolves the residual distributional inconsistencies inherent in parameter-level composition. Empirically, across five preference alignment benchmarks, \textsc{Mapes} attains the highest benchmark-level average for every backbone, with no large degradation on GSM8K and GPQA. Matched controls attribute the gains to their sources: uniform composition of the experts is already competitive, ordinary post-merge DPO recovers most of what a static merge loses, the preference direction and its per-coordinate weighting add a further, smaller gain, and the ensemble-KL term contributes mainly to honesty.

\section*{Limitations}
\label{sec:limitations}

\textsc{Mapes} has three principal limitations. First, the framework relies on preference datasets annotated with fine-grained value-specific scores. While UltraFeedback enables training on four core values (instruction-following, honesty, helpfulness, truthfulness), such annotations are expensive to obtain and not universally available. Extending \textsc{Mapes} to a broader value space, including dimensions such as fairness, harmlessness, or domain-specific safety constraints, would require either richer annotation pipelines or weaker forms of value supervision (e.g., synthetic value labels from strong judges). Our observed positive transfer to harmlessness (\S\ref{sec:positive_transfer}) is encouraging but is not a substitute for direct supervision over additional values.

Second, our experiments cover four instruction-tuned backbones of 3B to 8B parameters from two model families, two sizes per family; \textsc{Mapes} is the top method at each measured size (Appendix~\ref{app:scaling}), but the aggregate margin over the strongest baseline is smallest on the largest backbone (Llama-3.1-8B), and with two sizes per family we make no claim about the trend beyond the measured range. PAVM's per-value preference-direction estimation uses a fixed, small calibration set ($N{=}128$ per value), so this component does not grow with the training dataset; validating the pipeline on 13B+ and mixture-of-experts models remains future work. Relatedly, all compared methods live in the reward-model-free offline pairwise setting. PAVM's direction statistic is not algebraically specific to DPO (at the reference point it is the chosen-vs-rejected log-likelihood-gap gradient up to a positive scalar, a form shared by other pairwise objectives), but we have not established transfer to other alignment objectives empirically. Finally, the training pool shares prompts with TruthfulQA-MC for every compared method (Appendix~\ref{app:tqa_overlap}).

Third, our value conditioning is realized through prompt-based injection of scalar scores. While this design is simple and language-model-native, it may be sensitive to prompt formulation and does not guarantee fine-grained inference-time control over value trade-offs. More structured conditioning mechanisms, such as learned value-control tokens, compositional weight selection at the merging stage, or routing-style gating informed by user context, could further improve robustness and controllability of multi-value alignment systems built atop the \textsc{Mapes} paradigm. We view such developments as important future directions, building on the conflict-aware pipeline foundation introduced in this work.

\section*{Ethical Consideration}
\label{sec:ethical_consideration}
\textsc{Mapes} decomposes preference alignment into individual human values, enabling controllable analysis and reweighting of alignment objectives. While this design improves transparency, it also introduces risks if individual values are overemphasized in isolation, potentially leading to undesirable behaviors such as confident but inaccurate outputs.

Moreover, \textsc{Mapes} relies on value-annotated preference data, which may reflect subjective or culturally specific norms embedded in human judgments. As a result, value-specific behaviors learned by the model may not generalize uniformly across different social contexts.

We partially mitigate these risks through empirical analysis rather than direct deployment claims. Our conflict-aware merging study explicitly demonstrates failure modes of na\"ive value reweighting, highlighting the importance of balanced integration across values. \textsc{Mapes} is intended as a research framework for analyzing multi-value alignment, and practical applications would require additional safeguards and human oversight.

\bibliography{custom}

\clearpage
\appendix

\section{Training Dataset}
\label{app:training_dataset}

\begin{figure*}[t!]
    \centering
    \includegraphics[width=1.\linewidth]{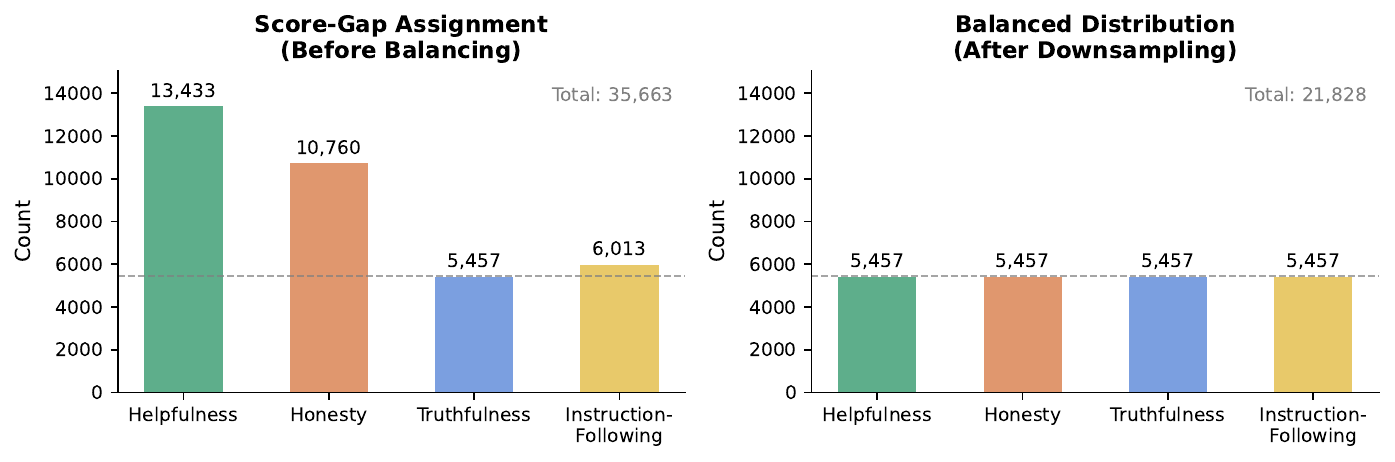}
    \caption{Distribution of data across the four values after score-gap routing. Left: the routed training partition (after the 2,000-pair held-out split) before balancing; right: the 21,828-pair training set after downsampling every value to the size of the smallest partition (5,457).}
    \label{fig:data_distribution}
\end{figure*}

\subsection{UltraFeedback Dataset}
\label{app:ultrafeedback_dataset}

To perform preference alignment tuning, we used the binarized UltraFeedback dataset.\footnote{Since the original UltraFeedback dataset is not inherently pairwise, we followed the standard binarization process previously used in related studies, available \href{https://huggingface.co/datasets/HuggingFaceH4/ultrafeedback_binarized}{here}.} We applied additional filtering to ensure data consistency. Specifically, we removed samples with missing value annotations, as well as those where the chosen response had a lower value score than the rejected response. We also excluded outlier samples with extremely long responses exceeding 6,000 tokens. After preprocessing, we retained 36k high-quality samples out of the original 55k.

\subsection{Data Preprocessing}
\label{app:data_preprocessing}

All reported results use the score-gap routing of \S\ref{sec:data_selection}; no source- or principle-based assignment is used. The procedure, with the counts produced by the released code, is: (1)~from the binarized UltraFeedback pairs, keep those with complete ratings for the four values, an overall preference in favor of the chosen response, and at most 6,000 tokens in total (43,080 pairs); (2)~assign each pair to the value with the largest positive score gap, discarding pairs with no positive gap; ties for the largest gap are broken by a fixed value order (helpfulness, honesty, truthfulness, instruction-following). This retains 37,663 pairs (helpfulness 14,169 / honesty 11,393 / instruction-following 6,352 / truthfulness 5,749; Figure~\ref{fig:data_distribution}, left); (3)~hold out 2,000 pairs for validation (random split, seed 42), used for checkpoint and learning-rate selection; (4)~downsample every value in the remaining training partition to the size of the smallest one, 5,457 pairs, giving 21,828 training pairs (Figure~\ref{fig:data_distribution}, right). The same 21,828 pairs are used by every method; only \textsc{Mapes} partitions them by value.

\subsection{Overlap with TruthfulQA-MC}
\label{app:tqa_overlap}

UltraFeedback contains prompts drawn from TruthfulQA. Matching the 817 TruthfulQA-MC validation questions against the prompts of the routed training pool (step~2 above, before balancing), 576 questions appear verbatim (747 within the full 55k binarized set). This affects every compared method identically, since all train on the same pairs, but it means the full TruthfulQA-MC set is not a clean held-out evaluation. On the 241 questions absent from the training pool, \textsc{Mapes} (Qwen2.5-7B) obtains MC1 44.81 / MC2 64.01, in line with its full-set scores (43.33 / 61.15). We release the list of overlapping question indices with the code.

\section{Prompt Design}
\label{app:prompt_exp}
As described in Section~\ref{sec:expert_training}, we train value expert adapters by conditioning them on target values and corresponding evaluation scores. These individual value-specific conditions are provided in the system prompt using structured natural language:

\begin{list}{}{\leftmargin=0.5em \rightmargin=0.5em}
    \item \texttt{You are an assistant that generates responses for the instruction while implicitly achieving the following target scores (on a scale of 1-5, where 1 is lowest and 5 is highest):}
    \item \texttt{[TARGET] score: [TARGET\_SCORE]}.
\end{list}

This approach is also applied during post-merge harmonization (\S\ref{sec:harmonization}). The key difference is that each value-specific adapter is trained only on the subset of data annotated with its corresponding value, whereas the unified adapter during harmonization is trained on the full set of individual value-annotated data across all values.

During inference, we use a multi-value mixture prompt to reflect all values concurrently. The system prompt includes all target scores as follows:

\begin{list}{}{\leftmargin=0.5em \rightmargin=0.5em}
    \item \texttt{You are an assistant that generates responses for the instruction while implicitly achieving the following target scores (on a scale of 1--5, where 1 is lowest and 5 is highest):}
    \item \texttt{Helpfulness score: [HELPFULNESS],}\\
    \texttt{Honesty score: [HONESTY],}\\
    \texttt{Truthfulness score: [TRUTHFULNESS],}\\
    \texttt{
        \makebox[0.6\linewidth][l]{Instruction-Following score:} \\
        \makebox[1.0\linewidth][r]{[INSTRUCTION\_FOLLOWING].}
    }
\end{list}

\section{Training Setup and Considerations}
\label{app:training_details}

\subsection{Unified Training Protocol Across Methods}
To ensure a fair comparison, we use the same instruction-tuned reference model across all experiments. All methods are trained for an equivalent number of steps corresponding to 4 epochs.
For CPO, we apply the CDPO training strategy with a shared reference model (denoted CDPO in result tables).
For RA-DPO, which augments the dataset to increase training diversity, we limit training to 2 epochs to match the total number of update steps.
For MODPO, we use the helpfulness dimension as the primary reward ($D_k$) with uniform weighting across objectives and train for 4 epochs.
\textsc{Mapes} allocates 3 epochs for value-specific expert training and 1 epoch for post-merge harmonization, totaling 4 epochs.
The hyperparameters used in all experiments are listed in Table~\ref{tab:hyperparameters}. For every method, including \textsc{Mapes}, the learning rate was selected from $\{1\mathrm{e}{-6}, 5\mathrm{e}{-6}, 1\mathrm{e}{-5}\}$ by validation loss on the 2,000-pair held-out split of Appendix~\ref{app:data_preprocessing}, with the same grid and rule for all methods; no evaluation benchmark was used for model or hyperparameter selection. The selected values are in Table~\ref{tab:selected_lr}. Training was conducted on two NVIDIA A5000 GPUs.

\subsection{Balancing Value-Specific Training}
After the downsampling of Appendix~\ref{app:data_preprocessing} the four value partitions are equal-sized (5,457 pairs each), so no class-level reweighting is needed during expert training; the only per-sample weighting is the gap-weighted DPO loss of \S\ref{sec:expert_training}.

\begin{table}[t!]
    \centering
    \resizebox{\columnwidth}{!}{
    \begin{tabular}{ll}
    \toprule
    \textbf{Hyperparameter} & \textbf{Value} \\
    \midrule

    \rowcolor[HTML]{E4E4E4}
    \multicolumn{2}{c}{Optimization} \\
    Optimizer & adamw\_torch \\
    LR scheduler & cosine \\
    Learning rate & \{ 1e-06, 5e-06, 1e-05 \} \\
    Warmup ratio & \{ 0.1, 0.05 \}\\

    \rowcolor[HTML]{E4E4E4}
    \multicolumn{2}{c}{Training Schedule} \\
    Epochs & 4 (total) \\
    Batch size & \{ 32, 64 \} \\

    \rowcolor[HTML]{E4E4E4}
    \multicolumn{2}{c}{DPO Specific} \\
    DPO beta & 0.01 \\
    Label smoothing & 0.0 \\
    Loss type & sigmoid \\

    \rowcolor[HTML]{E4E4E4}
    \multicolumn{2}{c}{LoRA (Adapter)} \\
    r & 16 \\
    $\alpha$ & 64 \\
    Dropout & 0.05 \\

    \rowcolor[HTML]{E4E4E4}
    \multicolumn{2}{c}{\textsc{Mapes} Specific} \\
    Gap-weight exclusivity $\gamma$ (\S\ref{sec:expert_training}) & 1.0 \\
    Harmonization KL weight $\lambda$ (\S\ref{sec:harmonization}) & 0.7 \\
    PAVM $\alpha$ / trim density $d$ & 1.0 / 0.2 \\
    Calibration size $N$ (per value) & 128 \\

    \bottomrule
    \end{tabular}
    }
    \caption{Training hyperparameters. Braces list the search grid; the learning rate is selected per method and backbone by held-out validation loss (Table~\ref{tab:selected_lr}), all other settings are shared by every method.}
    \label{tab:hyperparameters}
    \vspace{-2px}
\end{table}

Table~\ref{tab:selected_lr} consolidates the selected learning rate for every method--backbone pair.

\begin{table}[h!]
    \centering
    \small
    \setlength{\tabcolsep}{4pt}
    \resizebox{\columnwidth}{!}{
    \begin{tabular}{lcccc}
    \toprule
    Method & Q-3B & Q-7B & L-3B & L-8B \\
    \midrule
    DPO & 1e-6 & 1e-6 & 1e-6 & 5e-6 \\
    CDPO & 1e-5 & 5e-6 & 5e-6 & 5e-6 \\
    RA-DPO & 5e-6 & 5e-6 & 5e-6 & 5e-6 \\
    MODPO & 1e-6 & 1e-6 & 1e-6 & 1e-6 \\
    \textsc{Mapes} (expert / harm.) & 5e-6 / 5e-6 & 5e-6 / 5e-6 & 5e-6 / 1e-5 & 1e-5 / 1e-5 \\
    \bottomrule
    \end{tabular}
    }
    \caption{Selected learning rate per method and backbone (Q = Qwen2.5, L = Llama-3.2-3B/-3.1-8B). All runs use an effective batch size of 32 (per-device batch $\times$ gradient accumulation). For \textsc{Mapes}, the two values correspond to the expert-training and harmonization stages.}
    \label{tab:selected_lr}
\end{table}

\section{PAVM Implementation Details}
\label{app:pavm_details}

\paragraph{Preference direction estimation.}
For each value $v$, we estimate the preference improvement direction $\hat{p}_v$ as the negative gradient of the chosen-vs-rejected NLL gap on a calibration set $\mathcal{D}_v^{\mathrm{cal}}$ of $N{=}128$ preference pairs sampled from value $v$'s training partition:
\begin{multline*}
    \hat{p}_v \;=\; -\frac{1}{N} \!\!\!\sum_{(x, y^{+}, y^{-}) \in \mathcal{D}_v^{\mathrm{cal}}}\!\!\! \nabla_{\theta} \big[\mathrm{NLL}_\theta(y^{+} \mid x) \\
    {} - \mathrm{NLL}_\theta(y^{-} \mid x)\big].
\end{multline*}
The calibration pairs are drawn from the value's training partition, so they are not disjoint from expert training, but they are disjoint from the 2,000-pair held-out split (see Appendix~\ref{app:tqa_overlap} for the evaluation-set overlap that affects all methods). Gradients are computed with respect to the base-layer weights underlying each LoRA-target module, with all expert adapters disabled (so $\pi = \pi_{\mathrm{base}}$). For 7B+ models with limited GPU memory, we restrict gradient tracking to attention projection layers (\texttt{q\_proj}, \texttt{k\_proj}, \texttt{v\_proj}, \texttt{o\_proj}) and zero-fill MLP positions; this preserves the dominant directions of preference improvement while keeping memory usage tractable.

\paragraph{Merge hyperparameters.}
Phase 1 trimming density $d{=}0.2$. Phase 3 alignment weight $\alpha{=}1.0$ and per-coordinate normalization across experts ($\tilde{a}_{v,j} = a_{v,j} / \max_{v'} a_{v',j}$, with the $\max$ clamped from below by $10^{-10}$ to avoid division by zero at coordinates where every expert has $a_{v,j}{=}0$). Phase 2 ties (positions where the delta-sum is zero) fall back to a magnitude-weighted vote, $\mathrm{sign}(\sum_v p_{v,j} \cdot |p_{v,j}|)$. The denominator clamp in Eq.~(\ref{eq:pavm_merge}) handles positions where no expert agrees with the elected sign by setting $\Delta_{\mathrm{merged},j}{=}0$.

\section{Theoretical Justification of PAVM}
\label{app:pavm_theory}

We provide a formal account of why PAVM's first-order preference signal $p_v$, estimated at the base reference $\theta_{0} := \theta_{\mathrm{base}}$, supplies a locally informative directional cue for the merged adapter. The argument has two parts: a smoothness-based bound on the validity of the first-order surrogate (Proposition 1), and three local properties of the merged update of Eq.~(\ref{eq:pavm_merge}) itself (Facts 1--2 and Proposition 2). All results are local statements under explicit assumptions standard in first-order optimization and weight-space analysis~\citep{foret2021sam, wortsman2022modelsoups, frankle2020lmc, rame2023rewardedsoups}; none is a global optimality claim.

\paragraph{Setup.}
For each value $v \in \mathcal{V}$, let $\mathcal{L}^{(v)}: \mathbb{R}^{d} \!\to\! \mathbb{R}$ denote the value-specific DPO loss evaluated on a calibration set of preference pairs from value $v$. The \emph{preference improvement direction} at $\theta_{0}$ is
\begin{equation*}
    p_{v} \;:=\; -\nabla_{\theta} \mathcal{L}^{(v)}(\theta_{0}),
\end{equation*}
which, at $\pi_{\theta_0} = \pi_{\mathrm{ref}} = \pi_{\mathrm{base}}$, simplifies to $(\beta/2)\,[\nabla_{\theta}\log \pi_{\mathrm{base}}(y^{+}|x) - \nabla_{\theta}\log \pi_{\mathrm{base}}(y^{-}|x)]$ up to a positive scalar (cf.~Eq.~\eqref{eq:pref_dir}). Let $\Dm$ (shorthand for $\Delta_{\mathrm{merged}}$) denote the PAVM-merged delta of Eq.~\eqref{eq:pavm_merge}.

\paragraph{Assumptions.}
\begin{itemize}[itemsep=1pt, topsep=2pt]
    \item[\textbf{(A1)}] \textbf{Local $L$-smoothness.} Each $\mathcal{L}^{(v)}$ has $L$-Lipschitz gradient on the ball $\mathcal{B}(\theta_{0}, R)$ of radius $R$ around $\theta_0$: for all $\theta, \theta' \in \mathcal{B}(\theta_{0}, R)$,
    \begin{equation*}
        \|\nabla \mathcal{L}^{(v)}(\theta) - \nabla \mathcal{L}^{(v)}(\theta')\| \le L\,\|\theta - \theta'\|.
    \end{equation*}
    \item[\textbf{(A2)}] \textbf{Bounded merged step.} $\|\Dm\| \le R$.
\end{itemize}

\medskip
\noindent\textbf{Proposition 1 (First-order approximation bound).}
Under \textnormal{(A1)--(A2)},
\begin{equation*}
\begin{aligned}
\bigg|\sum_{v \in \mathcal{V}} &\big[\mathcal{L}^{(v)}(\theta_{0} + \Dm) - \mathcal{L}^{(v)}(\theta_{0})\big] \\
{} &+ \sum_{v \in \mathcal{V}} \langle p_{v},\, \Dm\rangle \bigg| \;\le\; \frac{|\mathcal{V}|\,L\,R^{2}}{2}.
\end{aligned}
\end{equation*}

\noindent\textit{Proof.}\hspace{0.4em}
A standard consequence of $L$-Lipschitz gradient~\citep[Lemma~1.2.3]{nesterov2004introductory} is the two-sided bound
\begin{multline*}
    \big|\mathcal{L}^{(v)}(\theta_{0} + \Delta) - \mathcal{L}^{(v)}(\theta_{0}) \\
    {} - \langle \nabla \mathcal{L}^{(v)}(\theta_{0}),\, \Delta\rangle\big| \le \tfrac{L}{2}\,\|\Delta\|^{2},
\end{multline*}
valid for all $\Delta$ with $\theta_{0} + \Delta \in \mathcal{B}(\theta_{0}, R)$. Substituting $\nabla \mathcal{L}^{(v)}(\theta_{0}) = -p_{v}$ and $\Delta = \Dm$, and summing over $v \in \mathcal{V}$ with the triangle inequality, yields the bound.
\medskip

\medskip
\noindent\textbf{Setup for the merged update.}
Fix a coordinate $j$ and write $\mathcal{V}^{\star} := \mathcal{V}_{j}^{\star}$ for the sign-consistent set of \S\ref{sec:pavm}, $N := |\mathcal{V}^{\star}|$, $s := s_j$ for the elected sign, $\delta_v := \tilde{\Delta}_{v,j}$ for the trimmed expert deltas (all of sign $s$, so $\delta_v = s|\delta_v|$), $\tilde{a}_v := \tilde{a}_{v,j} \in [0,1]$ for the normalized clipped alignment, $w_v := 1 + \alpha \tilde{a}_v$ and $W := \sum_{v \in \mathcal{V}^{\star}} w_v = N + \alpha \sum_v \tilde{a}_v$. PAVM's coordinate is $\Delta^{P} := \sum_v w_v \delta_v / W$ (Eq.~\ref{eq:pavm_merge}) and the uniform (TIES-style, $\alpha{=}0$) coordinate is $\Delta^{U} := \frac{1}{N}\sum_v \delta_v$.

\medskip
\noindent\textbf{Fact 1 (Eq.~\ref{eq:pavm_merge} is a weighted least-squares barycenter).}
$\Delta^{P} = \arg\min_{\Delta} \sum_{v \in \mathcal{V}^{\star}} w_v (\Delta - \delta_v)^{2}$.

\noindent\textit{Proof.}\hspace{0.4em}
The objective is a convex quadratic in the scalar $\Delta$; setting its derivative $2\sum_v w_v(\Delta - \delta_v)$ to zero gives $\Delta = \sum_v w_v \delta_v / W$.
\medskip

\noindent\textbf{Fact 2 (Exact decomposition and sign preservation).}
\begin{equation*}
    \Delta^{P} - \Delta^{U} \;=\; \frac{\alpha}{W}\sum_{v \in \mathcal{V}^{\star}} \tilde{a}_v(\delta_v - \Delta^{U}) \;=\; \frac{\alpha N}{W}\,\mathrm{Cov}_v(\tilde{a}_v, \delta_v),
\end{equation*}
where $\mathrm{Cov}_v$ is the empirical covariance over $v \in \mathcal{V}^{\star}$ with uniform weights. Moreover $\mathrm{sign}(\Delta^{P}) = \mathrm{sign}(\Delta^{U}) = s$: the preference weighting rescales the magnitude of the merged coordinate but never flips the elected sign.

\noindent\textit{Proof.}\hspace{0.4em}
$\sum_v w_v \delta_v = N\Delta^{U} + \alpha\sum_v \tilde{a}_v \delta_v$, hence $\Delta^{P} - \Delta^{U} = [N\Delta^{U} + \alpha\sum_v \tilde{a}_v\delta_v - W\Delta^{U}]/W = \frac{\alpha}{W}[\sum_v \tilde{a}_v \delta_v - (\sum_v \tilde{a}_v)\Delta^{U}] = \frac{\alpha}{W}\sum_v \tilde{a}_v(\delta_v - \Delta^{U})$. Since $\sum_v(\delta_v - \Delta^{U}) = 0$, this equals $\frac{\alpha}{W}\sum_v(\tilde{a}_v - \bar{a})(\delta_v - \Delta^{U}) = \frac{\alpha N}{W}\mathrm{Cov}_v(\tilde{a}_v,\delta_v)$. Sign preservation follows because $\Delta^{P}$ is a convex combination of the $\delta_v$, which all have sign $s$. The same convexity gives $|\Delta^{P}_j| \le \max_{v \in \mathcal{V}^{\star}_j} |\tilde{\Delta}_{v,j}|$ at every coordinate, hence $\|\Dm\|^{2} \le \sum_j \max_v \tilde{\Delta}_{v,j}^{2} \le \sum_v \|\tilde{\Delta}_v\|^{2}$: the merged step never exceeds the experts' collective step, so Assumption~(A2) holds with $R = (\sum_v \|\tilde{\Delta}_v\|^{2})^{1/2}$ without further assumption.
\medskip

\noindent\textbf{Proposition 2 (Merged-update dominance).}
If $\{\tilde{a}_v\}$ and $\{|\delta_v|\}$ are similarly ordered over $v \in \mathcal{V}^{\star}$ (in particular whenever every $v \in \mathcal{V}^{\star}$ has $\mathrm{sign}(p_{v,j}) = s$ and $|p_{v,j}|$ is the same for all of them, since then $\tilde{a}_v \propto |\delta_v|$), then
\begin{equation*}
    |\Delta^{P}| \;\ge\; |\Delta^{U}|,
\end{equation*}
with equality iff $\alpha = 0$ or one of $\{\tilde{a}_v\}$, $\{|\delta_v|\}$ is constant on $\mathcal{V}^{\star}$. If they are oppositely ordered, the inequality reverses.

\noindent\textit{Proof.}\hspace{0.4em}
Writing $\delta_v = s|\delta_v|$ in Fact~2 gives $|\Delta^{P}| - |\Delta^{U}| = \frac{\alpha N}{W}\mathrm{Cov}_v(\tilde{a}_v, |\delta_v|)$, and Chebyshev's sum inequality gives $\mathrm{Cov}_v(\tilde{a}_v,|\delta_v|) \ge 0$ under similar ordering (and $\le 0$ under opposite ordering).
\medskip

\noindent\textbf{Corollary (First-order aggregate gain of the merged update).}
Let $\bar{p}_j := \sum_v p_{v,j}$. The first-order proxy of Proposition~1 is $\sum_v \langle p_v, \Delta_{\mathrm{merged}}\rangle = \sum_j \Delta_{\mathrm{merged},j}\,\bar{p}_j$, and by Fact~2
\begin{equation*}
    \Delta^{P}_j \bar{p}_j - \Delta^{U}_j \bar{p}_j \;=\; s_j \bar{p}_j\,\frac{\alpha N_j}{W_j}\,\mathrm{Cov}_v(\tilde{a}_{v,j}, |\tilde{\Delta}_{v,j}|).
\end{equation*}
Hence, on every coordinate where the elected sign agrees with the aggregate preference direction ($s_j\bar{p}_j > 0$) and the ordering condition of Proposition~2 holds, PAVM's first-order gain is at least that of uniform aggregation; Proposition~1 bounds the gap between this proxy and the true aggregate loss change by $|\mathcal{V}| L R^{2}/2$.

\paragraph{Interpretation.}
Proposition~1 establishes that the first-order proxy $\sum_{v}\langle p_{v}, \Dm\rangle$ is a controlled surrogate for the true aggregate preference change, with approximation error bounded by $\mathcal{O}(R^{2})$. Fact~1 shows that Eq.~(\ref{eq:pavm_merge}) is not a heuristic combination but the exact solution of an explicit per-coordinate objective in which each expert's vote is weighted by the evidence that its update moves along its own preference direction. Fact~2 shows that, relative to uniform aggregation, PAVM only rescales magnitudes in the elected direction, by an amount proportional to the covariance between alignment evidence and delta size, never flips a sign, and never produces a step larger than the experts' collective step (which makes (A2) automatic). Proposition~2 and its corollary turn this into a conditional dominance statement about the merged update itself, replacing the earlier score-level statement.

\paragraph{Limitations of the bound.}
Three caveats warrant explicit mention. First, Assumption (A1) is an idealization: real LLM losses are non-convex and not globally $L$-smooth. We therefore interpret all statements as \emph{local} properties within a small neighborhood around the base reference $\theta_{0}$; this assumption is standard in first-order optimization analysis~\citep{foret2021sam, wortsman2022modelsoups} and aligns with prior work on linear mode connectivity~\citep{frankle2020lmc, neyshabur2020transferlearning} and parameter-efficient model merging~\citep{ilharco2023editing, rame2023rewardedsoups, ortizjimenez2023tangentspace}. Second, the dominance of Proposition~2 is conditional on the ordering of alignment scores and delta magnitudes at each coordinate, and the corollary additionally requires the elected sign to agree with $\bar{p}_j$; neither condition is enforced by PAVM, so no unconditional improvement guarantee follows. Third, the weighting only acts on coordinates where at least two experts survive trimming with the same sign ($N_j \ge 2$); where $N_j = 1$ the merged coordinate equals that expert's delta regardless of $w$. The empirical test of these properties is the $\alpha{=}0$ and random-direction rows of Table~\ref{tab:ablation_results}.

\section{Analysis Experiment Details}
\label{app:analysis_details}

\subsection{Ceiling Performance Analysis (\S\ref{sec:pavm_ceiling_comp})}
\label{app:ceiling_details}

To evaluate the ceiling performance of individual value-specific experts, we evaluate each expert adapter independently on its corresponding benchmark.
Specifically, after completing value-specific expert training (\S\ref{sec:expert_training}), we activate only a single expert adapter at a time and evaluate it on the benchmark most closely associated with its target value:
\textsc{Instruction-Following} is evaluated on Arena-Hard-Auto (v0.1)~\cite{arena-hard-2024};
\textsc{Honesty} and \textsc{Helpfulness} on HHH-Alignment~\cite{hhh-2021};
and \textsc{Truthfulness} on TruthfulQA (average of MC1 and MC2)~\cite{lin-etal-2022-truthfulqa}.

For the merged model (\textsc{Mapes}), we evaluate the final harmonized adapter on all four benchmarks simultaneously.
For DPO baseline comparison, we use the standard DPO model trained on the same data with identical hyperparameters.
All evaluations use the Qwen2.5-3B-Instruct base model with the same inference configuration (bf16 precision, greedy decoding for multiple-choice tasks, sampling for generation tasks).

The radar chart in Figure~\ref{fig:ceiling_performance} visualizes these results, where each axis corresponds to a value dimension and the radial distance indicates performance.

\subsection{KL Divergence Trajectory Analysis (\S\ref{sec:signal_maintenace})}
\label{app:kl_trajectory_details}

We track KL divergence at two stages of the \textsc{Mapes} pipeline to characterize how value-specific representations evolve.

\paragraph{Expert training (Figure~\ref{fig:kl_trajectory}, left).}
During value-specific expert training, we compute the pairwise KL divergence between all $\binom{|\mathcal{V}|}{2}$ expert pairs at regular checkpoint intervals (every 100 training steps).
For each pair of experts $(v_i, v_j)$, we estimate the symmetric KL divergence $\tfrac{1}{2}\big(D_{\mathrm{KL}}(\pi_{\theta_{v_i}} \| \pi_{\theta_{v_j}}) + D_{\mathrm{KL}}(\pi_{\theta_{v_j}} \| \pi_{\theta_{v_i}})\big)$ by computing the mean token-level KL on a held-out subset of 50 samples from the training data.
We report the average over all pairs at each checkpoint. Increasing pairwise KL indicates that experts are specializing toward distinct value-aligned distributions.

\begin{table}[t!]
    \centering
    \small
    \resizebox{\columnwidth}{!}{
    \begin{tabular}{lccc}
    \toprule
    Model & DPO & \textsc{Mapes} (expert + harm.) & Overhead \\
    \midrule
    Qwen2.5-3B & 21.4 & 17.7 + 7.3 = 25.0 & $+17\%$ \\
    Qwen2.5-7B & 33.9 & 28.0 + 10.9 = 38.9 & $+15\%$ \\
    Llama-3.2-3B & 20.1 & 16.8 + 7.2 = 24.0 & $+19\%$ \\
    Llama-3.1-8B & 36.9 & 30.6 + 11.9 = 42.5 & $+15\%$ \\
    \bottomrule
    \end{tabular}
    }
    \caption{Training cost in GPU-hours. \textsc{Mapes} requires approximately 15--19\% more GPU-hours than standard DPO at matched optimization steps.}
    \label{tab:compute_cost}
\end{table}

\begin{table}[t!]
    \centering
    \small
    \resizebox{\columnwidth}{!}{
    \begin{tabular}{lc}
    \toprule
    Stage & Peak per GPU (of 24\,GB) \\
    \midrule
    Standard DPO & 21.0\,GB \\
    \textsc{Mapes} value-expert training & 23.5\,GB \\
    PAVM direction estimation + merge & 19.4\,GB \\
    \bottomrule
    \end{tabular}
    }
    \caption{Peak GPU memory per stage on Qwen2.5-7B.}
    \label{tab:peak_memory}
\end{table}

\paragraph{Post-merge harmonization (Figure~\ref{fig:kl_trajectory}, right).}
After PAVM merging, we track the KL divergence between the merged (harmonized) model $\pi_{\theta_{\mathrm{merged}}}$ and each individual expert $\pi_{\theta_v}$ during the harmonization training.
We compute $D_{\mathrm{KL}}(\pi_{\theta_v} \| \pi_{\theta_{\mathrm{merged}}})$ for each value $v \in \mathcal{V}$ at every 50 training steps, using the same held-out sample set.
Decreasing KL indicates that the merged model is converging toward the expert distributions, i.e., recovering the specialized knowledge from each expert into a coherent joint distribution.
All KL computations are performed using the Qwen2.5-7B-Instruct model with bf16 precision.

\begin{table}[t!]
    \centering
    \small
    \resizebox{\columnwidth}{!}{
    \begin{tabular}{lccc}
    \toprule
    Model & GSM8K & GPQA & Avg. \\
    \midrule
    Qwen2.5-3B & 56.25 & 24.33 & 40.29 \\
    Qwen2.5-3B + DPO & 61.41 & 26.56 & 43.99 \\
    Qwen2.5-3B + CDPO & \textbf{62.25} & \textbf{27.23} & \textbf{44.74} \\
    Qwen2.5-3B + MODPO & 60.50 & \textbf{27.23} & 43.87 \\
    Qwen2.5-3B + RA-DPO & 57.01 & 26.56 & 41.79 \\
    Qwen2.5-3B + RS & 61.03 & 22.99 & 42.01 \\
    \rowcolor[HTML]{EAECFF}
    Qwen2.5-3B + \textsc{Mapes} & 55.19 & 24.78 & 39.99 \\
    \midrule
    Qwen2.5-7B & 58.38 & 25.89 & 42.14 \\
    Qwen2.5-7B + DPO & 75.66 & 28.57 & 52.12 \\
    Qwen2.5-7B + CDPO & 68.99 & 25.67 & 47.33 \\
    Qwen2.5-7B + MODPO & 65.20 & 21.21 & 43.21 \\
    Qwen2.5-7B + RA-DPO & 76.72 & \textbf{29.24} & 52.98 \\
    Qwen2.5-7B + RS & 66.41 & 23.66 & 45.04 \\
    \rowcolor[HTML]{EAECFF}
    Qwen2.5-7B + \textsc{Mapes} & \textbf{80.97} & 29.21 & \textbf{55.09} \\
    \midrule
    Llama-3.2-3B & 66.87 & 24.78 & 45.83 \\
    Llama-3.2-3B + DPO & \textbf{69.45} & 23.88 & 46.67 \\
    Llama-3.2-3B + CDPO & 64.82 & 22.99 & 43.91 \\
    Llama-3.2-3B + MODPO & 68.39 & \textbf{26.79} & \textbf{47.59} \\
    Llama-3.2-3B + RA-DPO & 64.75 & 24.11 & 44.43 \\
    Llama-3.2-3B + RS & 68.76 & 25.67 & 47.22 \\
    \rowcolor[HTML]{EAECFF}
    Llama-3.2-3B + \textsc{Mapes} & 67.25 & 26.34 & 46.80 \\
    \midrule
    Llama-3.1-8B & 68.61 & 22.99 & 45.80 \\
    Llama-3.1-8B + DPO & 72.48 & 25.70 & 49.09 \\
    Llama-3.1-8B + CDPO & 70.51 & 23.44 & 46.98 \\
    Llama-3.1-8B + MODPO & 73.01 & 21.21 & 47.11 \\
    Llama-3.1-8B + RA-DPO & 71.04 & \textbf{26.12} & 48.58 \\
    Llama-3.1-8B + RS & 70.28 & 24.78 & 47.53 \\
    \rowcolor[HTML]{EAECFF}
    Llama-3.1-8B + \textsc{Mapes} & \textbf{74.45} & \textbf{26.12} & \textbf{50.29} \\
    \bottomrule
    \end{tabular}
    }
    \caption{Full academic benchmark results for all four backbones, under the protocol of \S\ref{sec:academic_evaluation}.}
    \label{tab:academic_3b}
\end{table}

\section{Computational Cost}
\label{app:compute_cost}

We report the practical training cost of \textsc{Mapes} against standard DPO on our 2$\times$NVIDIA A5000 (24\,GB) setup, at matched optimization steps (e.g., 2{,}729 vs.\ 2{,}732 steps on Qwen2.5-7B). All stages run with ZeRO-3, CPU offloading, and LoRA.

Preference-direction estimation and the PAVM merge itself add only a small one-time cost (about 0.3 GPU-hours at $N{=}128$), which is independent of the training dataset size.

\section{Full Academic Benchmark Results}
\label{app:academic_3b}

Table~\ref{tab:academic_3b} reports the exact numbers underlying Figure~\ref{fig:academic_bars}, extended to all four backbones under the identical protocol (GSM8K 0-shot with regex-based answer extraction; GPQA log-likelihood scoring). The 3B backbones show the same pattern as 7B/8B on GSM8K (within about one point of the base model); on Qwen2.5-3B the two-test average is slightly below the base model (39.99 vs.\ 40.29), so we claim no large degradation rather than preservation. We note that the large GSM8K gain of CDPO on Qwen2.5-3B follows the same answer-format-compliance pattern discussed in \S\ref{sec:academic_evaluation}.

\section{Inference-Prompt Control Experiment}
\label{app:prompt_control}

Each method in Table~\ref{tab:main_results} is evaluated under its own native inference prompt. To rule out the possibility that \textsc{Mapes}'s advantage stems from its multi-value mixture prompt rather than from the learned policy, we re-evaluated the principal baselines under \textsc{Mapes}'s mixture prompt on Qwen2.5-7B. As shown in Table~\ref{tab:prompt_control}, switching DPO from its native prompt to the mixture prompt decreases HHH honesty (54.10 to 49.18), leaves HHH helpfulness nearly unchanged (42.41 to 42.37), and increases TruthfulQA MC1/MC2 (34.80/53.25 to 38.80/55.75). Nevertheless, this prompt transfer does not close the HHH honesty/helpfulness gap to \textsc{Mapes}. This suggests that the observed advantage is not explained by the inference prompt alone under the tested Qwen2.5-7B setting.

\begin{table}[t!]
    \centering
    \small
    \setlength{\tabcolsep}{4pt}
    \resizebox{\columnwidth}{!}{
    \begin{tabular}{lccc}
    \toprule
    Model & Honesty & Helpful. & MC1 / MC2 \\
    \midrule
    DPO (base prompt) & 54.10 & 42.37 & 34.80 / 53.25 \\
    DPO (mixture prompt) & 49.18 & 42.37 & 38.80 / 55.75 \\
    MODPO (mixture prompt) & 49.18 & 42.37 & 42.59 / 60.73 \\
    \rowcolor[HTML]{EAECFF}
    \textsc{Mapes} (mixture prompt) & \textbf{72.13} & \textbf{67.80} & \textbf{43.33} / \textbf{61.15} \\
    \bottomrule
    \end{tabular}
    }
    \caption{Prompt-control experiment on Qwen2.5-7B. Applying \textsc{Mapes}'s inference prompt to baselines does not close the gap, ruling out a prompt-conditioning confound.}
    \label{tab:prompt_control}
\end{table}

\section{PAVM Sensitivity Analysis}
\label{app:pavm_sensitivity}

We assess PAVM's robustness to its two estimation hyperparameters on Qwen2.5-7B: the trim density $d$ and the calibration-set size $N$, varying one axis at a time around the default $(d{=}0.2, N{=}128)$ and re-running the merge and harmonization for each setting (Table~\ref{tab:pavm_sensitivity}).

\begin{table}[h!]
    \centering
    \small
    \setlength{\tabcolsep}{4pt}
    \resizebox{\columnwidth}{!}{
    \begin{tabular}{lccccc}
    \toprule
    Variant & IFEval & Honesty & Helpful. & MC1 & MC2 \\
    \midrule
    $d{=}0.2$, $N{=}128$ (default) & 39.93 & 72.13 & 67.80 & 43.33 & 61.15 \\
    \midrule
    $d{=}0.1$ & 40.12 & 70.54 & 67.37 & 42.76 & 60.43 \\
    $d{=}0.3$ & 39.53 & 71.24 & 67.46 & 43.22 & 60.98 \\
    $d{=}0.5$ & 40.28 & 70.89 & 67.68 & 43.21 & 60.52 \\
    \midrule
    $N{=}32$ & 40.58 & 72.04 & 66.55 & 42.01 & 59.73 \\
    $N{=}64$ & 41.33 & 71.27 & 66.42 & 42.18 & 59.79 \\
    $N{=}256$ & 39.84 & 72.09 & 67.73 & 42.51 & 60.91 \\
    \bottomrule
    \end{tabular}
    }
    \caption{PAVM sensitivity to trim density $d$ and calibration size $N$ on Qwen2.5-7B under the final pipeline; the default row is the Table~\ref{tab:main_results} model. All variants remain within two points of the default on every metric, with no collapse on any axis.}
    \label{tab:pavm_sensitivity}
\end{table}

Across both axes, performance varies by at most about two points on every metric, indicating that PAVM is robust to the trim density and the calibration-set size. We additionally verified robustness to the direction-estimation reference point: estimating $p_v$ at the expert checkpoints instead of the base model leaves HHH honesty and helpfulness essentially unchanged.

\section{Harmonization Controls}
\label{app:harm_controls}

Table~\ref{tab:harm_controls} reports the matched controls for Stage~4 on Qwen2.5-7B. All rows start from the identical PAVM-merged checkpoint (the checkpoint-0 of the Table~\ref{tab:main_results} run) and train for the same 683 steps at effective batch size 32 with the same learning rate; only the KL weight $\lambda$ changes ($\lambda{=}0$ is DPO only). The $\lambda{=}0$ run uses micro-batch $4{\times}8$ and the $\lambda{=}0.35$ run $2{\times}16$ on one NVIDIA A6000, versus $1{\times}32$ for the paper run; the effective batch is unchanged. Arena-Hard for the $\lambda{=}0$ row is given in Table~\ref{tab:ablation_results}; it was not run for $\lambda{=}0.35$.

\begin{table}[t!]
    \centering
    \small
    \setlength{\tabcolsep}{3.5pt}
    \resizebox{\columnwidth}{!}{
    \begin{tabular}{lcccccc}
    \toprule
    Stage 4 variant & IFEval & Honest & Helpful & HH-RLHF & MC1 & MC2 \\
    \midrule
    none (PAVM merge only) & 19.96 & 52.46 & 45.76 & 63.30 & 36.47 & 55.46 \\
    DPO only ($\lambda{=}0$) & 34.01 & 68.85 & 69.49 & 61.64 & 44.80 & 61.50 \\
    DPO + KL, $\lambda{=}0.35$ & 31.42 & 72.13 & 67.80 & 61.77 & 44.92 & 62.20 \\
    DPO + KL, $\lambda{=}0.7$ (\textsc{Mapes}) & 39.93 & 72.13 & 67.80 & 63.22 & 43.33 & 61.15 \\
    \bottomrule
    \end{tabular}
    }
    \caption{Stage-4 controls on Qwen2.5-7B from the same merged checkpoint at matched steps. Post-merge DPO alone recovers most of the drop; the KL term gives a consistent gain on honesty and none on TruthfulQA. Six-column averages: 45.6 / 56.7 / 56.7 / 57.9.}
    \label{tab:harm_controls}
\end{table}

\section{RS Weight Sweep}
\label{app:rs_sweep}

RS in Table~\ref{tab:main_results} is the uniform soup of the four value experts with no further training. Table~\ref{tab:rs_sweep} varies the interpolation weights on Qwen2.5-7B (one value at 0.4, the others at 0.2), again without training. No weighting closes the gap to \textsc{Mapes}; the benchmark-level macro average over the four benchmarks reported here (Arena-Hard excluded) ranges from 51.1 to 53.7, versus 53.4 for the uniform soup and 56.3 for \textsc{Mapes}. Because RS performs no post-merge updates, its training budget is three quarters of that of every other method in Table~\ref{tab:main_results}.

\begin{table}[h!]
    \centering
    \small
    \setlength{\tabcolsep}{3.5pt}
    \resizebox{\columnwidth}{!}{
    \begin{tabular}{lcccccc}
    \toprule
    Weights (help, hon, truth, IF) & IFEval & Honest & Helpful & HH-RLHF & MC1 & MC2 \\
    \midrule
    uniform (RS) & 34.94 & 67.21 & 59.32 & 63.04 & 45.04 & 60.03 \\
    (0.4, 0.2, 0.2, 0.2) & 32.16 & 65.57 & 57.63 & 62.62 & 40.02 & 55.74 \\
    (0.2, 0.4, 0.2, 0.2) & 32.35 & 65.57 & 64.41 & 63.17 & 47.12 & 61.56 \\
    (0.2, 0.2, 0.4, 0.2) & 31.98 & 65.57 & 61.02 & 63.38 & 47.61 & 61.87 \\
    (0.2, 0.2, 0.2, 0.4) & 32.90 & 63.93 & 57.63 & 63.08 & 42.84 & 58.67 \\
    \rowcolor[HTML]{EAECFF}
    \textsc{Mapes} & 39.93 & 72.13 & 67.80 & 63.22 & 43.33 & 61.15 \\
    \bottomrule
    \end{tabular}
    }
    \caption{Non-uniform soups of the same experts on Qwen2.5-7B (no post-merge training).}
    \label{tab:rs_sweep}
\end{table}

\section{Gradient Geometry at the Base Model}
\label{app:grad_geometry}

\begin{table}[h!]
    \centering
    \small
    \begin{tabular}{lc}
    \toprule
    Value pair & Gradient cosine \\
    \midrule
    helpfulness $\leftrightarrow$ honesty & $+0.61$ \\
    honesty $\leftrightarrow$ truthfulness & $+0.55$ \\
    helpfulness $\leftrightarrow$ truthfulness & $+0.54$ \\
    honesty $\leftrightarrow$ instruction-following & $+0.50$ \\
    truthfulness $\leftrightarrow$ instruction-following & $+0.49$ \\
    helpfulness $\leftrightarrow$ instruction-following & $+0.39$ \\
    \midrule
    Mean (all pairs) & $+0.51$ \\
    \bottomrule
    \end{tabular}
    \caption{Pairwise cosine similarity between per-value preference-improvement gradients, over the attention-projection coordinates used by PAVM.}
    \label{tab:grad_conflict}
\end{table}

Table~\ref{tab:grad_conflict} gives the pairwise cosine similarities discussed in \S\ref{sec:grad_conflict}. All six are positive, so the summed preference gradient has a positive inner product with every per-value gradient in the measured subspace; the statistics describe moderate alignment with frequent coordinate-level sign disagreement, not an optimization conflict in which improving one value must hurt another.

\section{Minimal Recipe}
\label{app:recipe}

\textsc{Mapes} reduces to four commands on top of a standard DPO trainer: (1)~route each preference pair to the value with the largest positive score gap and balance the partitions (Appendix~\ref{app:data_preprocessing}); (2)~train one LoRA adapter per value with DPO on its partition (3 epochs); (3)~run the merge script, which estimates $p_v$ from $N{=}128$ pairs per value and applies Eq.~(\ref{eq:pavm_merge}) (about 0.3 GPU-hours); (4)~train the merged adapter for one epoch with Eq.~(\ref{eq:harm}). No reward model or additional data is required, the total cost is 15--19\% above standard DPO (Appendix~\ref{app:compute_cost}), the \textsc{Mapes}-specific hyperparameters ($\gamma$, $\lambda$, $\alpha$, $d$, $N$) are fixed across backbones, and the output is a single adapter with no inference-time overhead.

\section{Results Across Model Sizes}
\label{app:scaling}

\begin{figure}[h!]
    \centering
    \includegraphics[width=\linewidth]{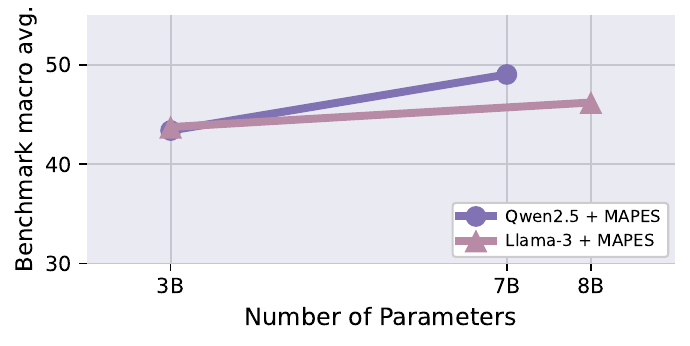}
    \caption{\textsc{Mapes} benchmark-level macro average at the two measured sizes per family (measured points only; no fitted trend).}
    \label{fig:scalability}
\end{figure}

\textsc{Mapes} is the top method at each of the four measured sizes (Table~\ref{tab:main_results}). Its absolute gain over the instruction-tuned base model on the benchmark-level average is larger at the larger size in both families ($+6.9 \!\to\! +7.8$ points on Qwen2.5, $+9.3 \!\to\! +12.0$ on Llama-3), while its margin over the strongest baseline is smallest on Llama-3.1-8B. With two sizes per family we do not fit or extrapolate a trend.

\section{Per-Benchmark Seed Variability}
\label{app:seed_std}

Table~\ref{tab:seed_std} reports the per-cell mean and sample standard deviation over training seeds 42/43/44 for all methods and backbones in Table~\ref{tab:main_results}. Per-benchmark deviations are mostly below one point, and the standard deviation of the five-benchmark macro average is below 0.8 for every method. However, some deviations are non-negligible relative to the smallest between-method margins, so mean differences should be interpreted together with seed-level variability.

\begin{table*}[t!]
    \centering
    \small
    \setlength{\tabcolsep}{3pt}
    \resizebox{\textwidth}{!}{
    \begin{tabular}{lccccccccc}
    \toprule
    Model & Arena v0.1 & Arena v2.0 & IFEval & HHH-Hon. & HHH-Help. & HH-RLHF & MC1 & MC2 & Avg. \\
    \midrule
    Qwen2.5-3B & 19.82{\scriptsize$\pm$0.50} & 2.40{\scriptsize$\pm$0.65} & 35.92{\scriptsize$\pm$0.46} & 44.26{\scriptsize$\pm$0.00} & 38.98{\scriptsize$\pm$0.00} & 60.83{\scriptsize$\pm$0.00} & 24.72{\scriptsize$\pm$0.00} & 41.41{\scriptsize$\pm$0.00} & 36.51{\scriptsize$\pm$0.05} \\
    Qwen2.5-3B + DPO & 19.64{\scriptsize$\pm$0.62} & 2.22{\scriptsize$\pm$0.81} & 35.30{\scriptsize$\pm$0.56} & 44.26{\scriptsize$\pm$3.28} & 44.07{\scriptsize$\pm$1.69} & 61.21{\scriptsize$\pm$0.30} & 30.35{\scriptsize$\pm$0.53} & 47.45{\scriptsize$\pm$0.52} & 38.10{\scriptsize$\pm$0.35} \\
    Qwen2.5-3B + CDPO & 21.12{\scriptsize$\pm$0.57} & 1.84{\scriptsize$\pm$0.55} & 35.92{\scriptsize$\pm$0.77} & 45.36{\scriptsize$\pm$0.95} & 42.37{\scriptsize$\pm$1.69} & 61.04{\scriptsize$\pm$0.24} & 37.33{\scriptsize$\pm$0.56} & 56.24{\scriptsize$\pm$0.59} & 39.82{\scriptsize$\pm$0.33} \\
    Qwen2.5-3B + MODPO & 19.90{\scriptsize$\pm$0.58} & 2.50{\scriptsize$\pm$0.62} & 35.67{\scriptsize$\pm$0.37} & 45.90{\scriptsize$\pm$2.84} & 40.68{\scriptsize$\pm$1.69} & 62.70{\scriptsize$\pm$0.26} & 33.91{\scriptsize$\pm$0.68} & 49.93{\scriptsize$\pm$1.24} & 38.96{\scriptsize$\pm$0.13} \\
    Qwen2.5-3B + RA-DPO & 22.52{\scriptsize$\pm$0.37} & 2.74{\scriptsize$\pm$0.40} & 35.43{\scriptsize$\pm$1.06} & 47.54{\scriptsize$\pm$1.64} & 48.59{\scriptsize$\pm$0.98} & 62.72{\scriptsize$\pm$0.39} & 27.21{\scriptsize$\pm$0.49} & 47.12{\scriptsize$\pm$0.54} & 39.20{\scriptsize$\pm$0.17} \\
    Qwen2.5-3B + RS & 21.80{\scriptsize$\pm$0.45} & 3.20{\scriptsize$\pm$0.80} & 34.57{\scriptsize$\pm$0.55} & 47.54{\scriptsize$\pm$1.64} & 40.68{\scriptsize$\pm$3.39} & 61.55{\scriptsize$\pm$0.40} & 33.41{\scriptsize$\pm$0.65} & 53.09{\scriptsize$\pm$0.70} & 39.20{\scriptsize$\pm$0.18} \\
    Qwen2.5-3B + \textsc{Mapes} & 19.54{\scriptsize$\pm$0.38} & 2.86{\scriptsize$\pm$0.85} & 35.92{\scriptsize$\pm$0.57} & 57.38{\scriptsize$\pm$1.64} & 49.15{\scriptsize$\pm$1.69} & 61.32{\scriptsize$\pm$0.64} & 47.00{\scriptsize$\pm$0.74} & 63.38{\scriptsize$\pm$0.18} & 43.38{\scriptsize$\pm$0.27} \\
    \midrule
    Qwen2.5-7B & 29.80{\scriptsize$\pm$0.99} & 3.72{\scriptsize$\pm$0.55} & 39.74{\scriptsize$\pm$0.49} & 49.18{\scriptsize$\pm$0.00} & 38.98{\scriptsize$\pm$0.00} & 62.66{\scriptsize$\pm$0.00} & 33.66{\scriptsize$\pm$0.00} & 51.64{\scriptsize$\pm$0.00} & 41.18{\scriptsize$\pm$0.06} \\
    Qwen2.5-7B + DPO & 31.14{\scriptsize$\pm$0.43} & 3.54{\scriptsize$\pm$1.07} & 39.25{\scriptsize$\pm$0.57} & 54.10{\scriptsize$\pm$1.64} & 42.37{\scriptsize$\pm$3.39} & 62.53{\scriptsize$\pm$0.23} & 34.80{\scriptsize$\pm$1.13} & 53.25{\scriptsize$\pm$0.39} & 42.28{\scriptsize$\pm$0.48} \\
    Qwen2.5-7B + CDPO & 31.42{\scriptsize$\pm$0.42} & 4.26{\scriptsize$\pm$0.75} & 38.32{\scriptsize$\pm$0.39} & 55.74{\scriptsize$\pm$1.64} & 52.54{\scriptsize$\pm$3.39} & 61.12{\scriptsize$\pm$1.09} & 38.92{\scriptsize$\pm$0.65} & 60.34{\scriptsize$\pm$0.63} & 44.21{\scriptsize$\pm$0.43} \\
    Qwen2.5-7B + MODPO & 29.60{\scriptsize$\pm$1.12} & 4.00{\scriptsize$\pm$0.65} & 40.48{\scriptsize$\pm$0.67} & 50.82{\scriptsize$\pm$1.64} & 44.07{\scriptsize$\pm$1.69} & 62.34{\scriptsize$\pm$0.35} & 36.84{\scriptsize$\pm$0.86} & 55.56{\scriptsize$\pm$0.34} & 42.65{\scriptsize$\pm$0.46} \\
    Qwen2.5-7B + RA-DPO & 31.54{\scriptsize$\pm$0.52} & 4.82{\scriptsize$\pm$0.47} & 39.50{\scriptsize$\pm$0.53} & 62.30{\scriptsize$\pm$1.64} & 58.76{\scriptsize$\pm$0.98} & 62.41{\scriptsize$\pm$0.63} & 38.84{\scriptsize$\pm$0.60} & 60.71{\scriptsize$\pm$0.84} & 46.08{\scriptsize$\pm$0.33} \\
    Qwen2.5-7B + RS & 31.20{\scriptsize$\pm$0.52} & 4.80{\scriptsize$\pm$0.58} & 34.94{\scriptsize$\pm$0.50} & 67.21{\scriptsize$\pm$3.28} & 59.32{\scriptsize$\pm$2.94} & 63.04{\scriptsize$\pm$0.42} & 45.04{\scriptsize$\pm$0.71} & 60.03{\scriptsize$\pm$0.60} & 46.36{\scriptsize$\pm$0.21} \\
    Qwen2.5-7B + \textsc{Mapes} & 34.20{\scriptsize$\pm$0.42} & 5.28{\scriptsize$\pm$0.28} & 39.93{\scriptsize$\pm$0.49} & 72.13{\scriptsize$\pm$1.64} & 67.80{\scriptsize$\pm$1.69} & 63.22{\scriptsize$\pm$0.62} & 43.33{\scriptsize$\pm$0.98} & 61.15{\scriptsize$\pm$0.49} & 49.02{\scriptsize$\pm$0.21} \\
    \midrule
    Llama-3.2-3B & 10.90{\scriptsize$\pm$0.65} & 1.02{\scriptsize$\pm$0.68} & 36.60{\scriptsize$\pm$0.66} & 42.62{\scriptsize$\pm$0.00} & 38.98{\scriptsize$\pm$0.00} & 53.65{\scriptsize$\pm$0.00} & 24.24{\scriptsize$\pm$0.00} & 46.32{\scriptsize$\pm$0.00} & 34.46{\scriptsize$\pm$0.10} \\
    Llama-3.2-3B + DPO & 11.24{\scriptsize$\pm$0.49} & 1.10{\scriptsize$\pm$0.46} & 36.04{\scriptsize$\pm$0.37} & 49.18{\scriptsize$\pm$1.64} & 40.68{\scriptsize$\pm$1.69} & 61.82{\scriptsize$\pm$0.42} & 37.13{\scriptsize$\pm$0.55} & 57.51{\scriptsize$\pm$0.52} & 39.26{\scriptsize$\pm$0.09} \\
    Llama-3.2-3B + CDPO & 10.24{\scriptsize$\pm$0.42} & 0.84{\scriptsize$\pm$0.56} & 35.31{\scriptsize$\pm$0.96} & 57.38{\scriptsize$\pm$1.64} & 52.54{\scriptsize$\pm$2.94} & 62.61{\scriptsize$\pm$0.64} & 40.31{\scriptsize$\pm$0.72} & 60.05{\scriptsize$\pm$1.07} & 41.72{\scriptsize$\pm$0.26} \\
    Llama-3.2-3B + MODPO & 12.90{\scriptsize$\pm$0.56} & 0.80{\scriptsize$\pm$0.08} & 36.66{\scriptsize$\pm$0.46} & 45.90{\scriptsize$\pm$1.64} & 44.07{\scriptsize$\pm$1.69} & 61.38{\scriptsize$\pm$0.22} & 39.29{\scriptsize$\pm$0.97} & 58.00{\scriptsize$\pm$0.24} & 39.70{\scriptsize$\pm$0.41} \\
    Llama-3.2-3B + RA-DPO & 9.02{\scriptsize$\pm$0.77} & 0.94{\scriptsize$\pm$0.28} & 36.41{\scriptsize$\pm$0.49} & 59.02{\scriptsize$\pm$2.84} & 49.15{\scriptsize$\pm$1.69} & 62.33{\scriptsize$\pm$0.32} & 43.13{\scriptsize$\pm$0.60} & 60.72{\scriptsize$\pm$0.30} & 41.95{\scriptsize$\pm$0.36} \\
    Llama-3.2-3B + RS & 11.10{\scriptsize$\pm$0.49} & 1.10{\scriptsize$\pm$0.36} & 33.27{\scriptsize$\pm$0.54} & 50.82{\scriptsize$\pm$1.64} & 44.07{\scriptsize$\pm$2.94} & 64.10{\scriptsize$\pm$0.43} & 40.27{\scriptsize$\pm$0.66} & 59.63{\scriptsize$\pm$0.52} & 40.17{\scriptsize$\pm$0.19} \\
    Llama-3.2-3B + \textsc{Mapes} & 11.64{\scriptsize$\pm$0.58} & 1.32{\scriptsize$\pm$0.37} & 34.38{\scriptsize$\pm$0.56} & 62.30{\scriptsize$\pm$1.64} & 62.71{\scriptsize$\pm$1.69} & 62.32{\scriptsize$\pm$0.44} & 44.80{\scriptsize$\pm$0.56} & 61.02{\scriptsize$\pm$0.60} & 43.72{\scriptsize$\pm$0.21} \\
    \midrule
    Llama-3.1-8B & 15.22{\scriptsize$\pm$0.43} & 1.64{\scriptsize$\pm$0.47} & 28.53{\scriptsize$\pm$0.46} & 44.26{\scriptsize$\pm$0.00} & 38.98{\scriptsize$\pm$0.00} & 55.18{\scriptsize$\pm$0.00} & 26.32{\scriptsize$\pm$0.00} & 48.51{\scriptsize$\pm$0.00} & 34.24{\scriptsize$\pm$0.05} \\
    Llama-3.1-8B + DPO & 16.94{\scriptsize$\pm$0.55} & 1.68{\scriptsize$\pm$0.36} & 33.64{\scriptsize$\pm$0.37} & 62.30{\scriptsize$\pm$1.64} & 55.93{\scriptsize$\pm$2.94} & 63.20{\scriptsize$\pm$0.54} & 39.49{\scriptsize$\pm$0.14} & 58.54{\scriptsize$\pm$0.74} & 42.86{\scriptsize$\pm$0.42} \\
    Llama-3.1-8B + CDPO & 18.16{\scriptsize$\pm$0.54} & 1.72{\scriptsize$\pm$0.19} & 33.33{\scriptsize$\pm$0.65} & 60.66{\scriptsize$\pm$3.28} & 57.63{\scriptsize$\pm$1.69} & 65.04{\scriptsize$\pm$0.87} & 46.02{\scriptsize$\pm$0.68} & 62.23{\scriptsize$\pm$0.98} & 44.32{\scriptsize$\pm$0.73} \\
    Llama-3.1-8B + MODPO & 18.20{\scriptsize$\pm$0.29} & 1.40{\scriptsize$\pm$0.70} & 36.60{\scriptsize$\pm$0.98} & 47.54{\scriptsize$\pm$1.64} & 48.59{\scriptsize$\pm$0.98} & 63.51{\scriptsize$\pm$0.23} & 38.19{\scriptsize$\pm$0.37} & 56.91{\scriptsize$\pm$0.68} & 41.10{\scriptsize$\pm$0.13} \\
    Llama-3.1-8B + RA-DPO & 16.64{\scriptsize$\pm$0.64} & 1.42{\scriptsize$\pm$0.70} & 34.20{\scriptsize$\pm$0.64} & 59.02{\scriptsize$\pm$1.64} & 55.93{\scriptsize$\pm$1.69} & 65.22{\scriptsize$\pm$0.63} & 45.00{\scriptsize$\pm$0.56} & 62.24{\scriptsize$\pm$0.56} & 43.91{\scriptsize$\pm$0.15} \\
    Llama-3.1-8B + RS & 20.60{\scriptsize$\pm$0.47} & 2.40{\scriptsize$\pm$0.56} & 35.67{\scriptsize$\pm$0.48} & 59.02{\scriptsize$\pm$1.64} & 52.54{\scriptsize$\pm$1.69} & 64.40{\scriptsize$\pm$0.41} & 46.63{\scriptsize$\pm$0.64} & 63.47{\scriptsize$\pm$0.72} & 44.48{\scriptsize$\pm$0.20} \\
    Llama-3.1-8B + \textsc{Mapes} & 19.24{\scriptsize$\pm$0.56} & 2.16{\scriptsize$\pm$0.87} & 37.03{\scriptsize$\pm$0.28} & 63.93{\scriptsize$\pm$1.64} & 61.02{\scriptsize$\pm$1.69} & 65.22{\scriptsize$\pm$0.50} & 46.43{\scriptsize$\pm$0.35} & 64.72{\scriptsize$\pm$1.00} & 46.20{\scriptsize$\pm$0.51} \\
    \bottomrule
    \end{tabular}
    }
    \caption{Per-benchmark mean and sample standard deviation over training seeds 42/43/44, for all methods and backbones in Table~\ref{tab:main_results}.}
    \label{tab:seed_std}
\end{table*}

\end{CJK}

\end{document}